\documentclass[lettersize,journal]{IEEEtran}
\usepackage{amsmath,amsfonts}
\usepackage{algorithmic}
\usepackage{algorithm}
\usepackage{array}
\usepackage[caption=false,font=normalsize,labelfont=sf,textfont=sf]{subfig}
\usepackage{textcomp}
\usepackage{stfloats}
\usepackage{url}
\usepackage{verbatim}
\usepackage{graphicx}
\usepackage{cite}
\usepackage{xcolor}
\usepackage{bbm}
\begin{document}

\title{Remote Heart Rate Monitoring in Smart Environments from Videos with\\Self-supervised Pre-training}

\author{\IEEEauthorblockN{~}
\IEEEauthorblockA{Divij Gupta, Ali Etemad\\Dept. ECE \& Ingenuity Labs Research Institute\\Queen's University, Canada\\
\{gupta.d,~ali.etemad\}@queensu.ca}
\thanks{Copyright (c) 2023 IEEE. Personal use of this material is permitted. However, permission to use this material for any other purposes must be obtained from the IEEE by sending a request to pubs-permissions@ieee.org}
}

\markboth {IEEE Internet of Things Journal}
{Shell \MakeLowercase{\textit{et al.}}: A Sample Article Using IEEEtran.cls for IEEE Journals}

\maketitle

\begin{abstract}
Recent advances in deep learning have made it increasingly feasible to estimate heart rate remotely in smart environments by analyzing videos. However, a notable limitation of deep learning methods is their heavy reliance on extensive sets of labeled data for effective training. To address this issue, self-supervised learning has emerged as a promising avenue. Building on this, we introduce a solution that utilizes self-supervised contrastive learning for the estimation of remote photoplethysmography (PPG) and heart rate monitoring, thereby reducing the dependence on labeled data and enhancing performance. We propose the use of 3 spatial and 3 temporal augmentations for training an encoder through a contrastive framework, followed by utilizing the late-intermediate embeddings of the encoder for remote PPG and heart rate estimation. Our experiments on two publicly available datasets showcase the improvement of our proposed approach over several related works as well as supervised learning baselines, as our results approach the state-of-the-art. We also perform thorough experiments to showcase the effects of using different design choices such as the video representation learning method, the augmentations used in the pre-training stage, and others. We also demonstrate the robustness of our proposed method over the supervised learning approaches on reduced amounts of labeled data.  
\end{abstract}


\section{Introduction}
\label{sec:intro}

Photoplethysmography (PPG) is an optical measurement that indicates the changes in blood volume. It is a relatively cheap and non-invasive method that uses a light source and detector to measure the change in light variation caused by blood flow through the flesh \cite{ppg}. A variety of different types of information is carried by or can be derived from PPG signals \cite{ppghaemo,ppgresp}, including hemoglobin levels, cardiovascular conditions, heart rate (HR), cardiac output, blood pressure, oxygen saturation level (SpO2), and even a subject's respiration rate. The signals have been used in a variety of non-medical applications as well, for instance in emotion recognition \cite{ppgemotion}, cognitive load assessment \cite{ppgload}, and others. 

\begin{figure}[t]
    \centering
    \includegraphics[width=0.90\columnwidth]{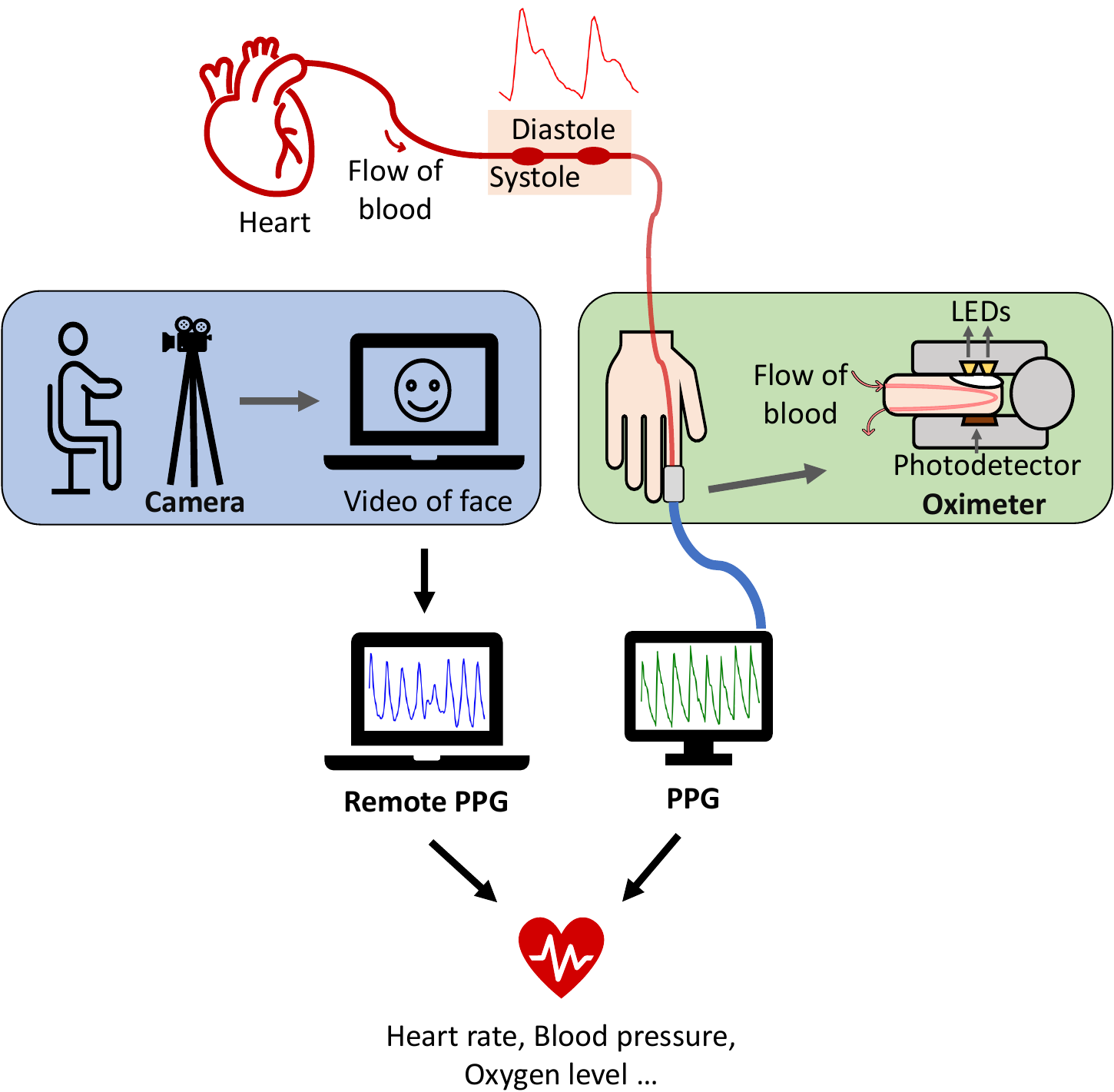}
    \caption{An overview of PPG collection using an Oximeter, as well as rPPG estimation using a standard video camera. In a typical contact-based oximeter, infrared light is passed through the finger and received by a photodetector. The change in the light received is an indicator of the blood flow and is used to measure PPG. For rPPG, the change in color on the surface of the skin is recorded by a camera and further processed to estimate the signal.}
    \label{fig:over}
\end{figure}

While PPG is conventionally measured through an oximeter worn by the user on a finger, studies have shown that blood flow, and consequently PPG, can also be measured from afar \cite{origin}, as blood flow causes subtle color variations at the surface of the skin. This process, termed remote PPG (rPPG) eliminates all forms of contact while giving the same benefits as a PPG signal acquired through an oximeter. The estimated rPPG, in turn, can be used to analyze cardiac activity, most notably by calculating HR values. So much so that the comparison between HR values derived from rPPG (estimated from the videos) and the PPG reference/ground-truth signals is often used as the main performance measure for rPPG algorithms. 
Hence, this is very useful in scenarios such as pandemics or virtual settings where direct access to the skin is not advised or always possible. Furthermore, since rPPG requires only a camera, it is very easy to integrate into existing Internet of Things (IoT)-enabled smart environments that 
comprise cameras, data transmission channels, and cloud servers for storing and processing information \cite{iotgen}. The various vitals and information that can be extracted from a PPG signal and the non-contact remote acquisition of rPPG provide a strong motivation for rPPG to be incorporated into smart homes, workplaces, hospitals, and others \cite{ppgiot,ppgiot2}. An overview of PPG and rPPG estimation is depicted in Figure \ref{fig:over}. Nonetheless, despite the numerous advantages of using rPPG instead of PPG, accurate and robust estimation of rPPG remains highly challenging due to factors such as visual noise, low spatiotemporal video resolution, improper illumination, varying skin tone, and others.

In recent years, rPPG estimation has become more robust as a result of advances in computer vision and deep learning \cite{HRCNN, deep}. A major limitation of supervised deep learning solutions is the reliance on huge amounts of annotated data for proper training. To address this, self-supervised learning has lately begun to gain momentum in the field of deep learning. The central concept behind this paradigm is to generate pseudo-labels instead of human-annotated labels, which would then be used to train the model. These pseudo-labels are often derived by performing various augmentations (transformations) on the available data. The model is then trained to recognize these augmentations, for instance, by detecting that two different transformations applied to the same input sample are indeed renditions of the same information. This will allow the network to learn to extract informative representations from the input data without requiring the actual output class labels. Following the self-supervised learning step, fine-tuning is often applied to train specific layers of the network for the downstream task. 

In this work, to provide an effective approach for rPPG estimation while reducing reliance on output labels, we propose a deep learning solution that leverages contrastive self-supervised pre-training. We believe, given the scarcity of datasets in the field of rPPG, our method provides a valuable avenue for tackling this problem. Our model uses a 3D convolution-based encoder to obtain representations of facial videos through self-supervised contrastive learning. The model is then fine-tuned for rPPG signal estimation, achieving strong results. Our contributions can be summarized as follows:
\begin{itemize}
    \item  We propose an effective solution for rPPG estimation in IoT settings, based on contrastive learning.
    
    \item  We perform thorough experiments on two publicly available datasets and validate the effectiveness of our method, showing that our solution achieves strong results in measuring rPPG and estimating HR without the need for contact-based sensors.
    
    \item We perform a large number of experiments to evaluate the impact of different design choices of the proposed method such as the pairing strategy and the augmentations used in the self-supervised training paradigm, the video representation learning technique, and the facial regions taken for extracting the rPPG. Further experiments demonstrate that our solution performs robustly when the amount of labeled data for training is reduced.

\end{itemize}

\section{Related Work}
In this section we first review prior works on rPPG estimation, followed by general self-supervised representation learning.

\subsection{rPPG Estimation}
\label{related}
A number of classical image processing methods have used color space transformations and signal processing approaches to estimate rPPG. In CHROM \cite{chrom}, the average intensity of skin pixels was computed for each frame of the video. These red, green, and blue (RGB) channel mean intensities were then tracked temporally across the frames to obtain 3 traces, one for each color channel. These traces were then bandpass filtered and combined linearly to obtain rPPG. A similar approach was taken in POS \cite{pos} where the RGB traces were projected onto an orthogonal color space to estimate the rPPG signal. In 2SR \cite{ssr}, a slightly different approach was used wherein the skin pixels were detected and a subspace of the skin pixels was created for each frame. Next, the temporal rotation across the subspaces was tracked to estimate the rPPG signals. A key difference in 2SR with respect to other works was the use of the spatial distribution of the skin pixels which was discarded in other classical methods since they used the average intensity values of the skin pixels in a frame. 

An interesting approach was taken in \cite{mit} where Eulerian video magnification was proposed. The authors used spatial decomposition, temporal filtering, and spatial reconstruction to amplify both the color as well as low-amplitude motion in the video. Since rPPG estimation primarily relies on the color variations on the skin surface, using color magnification made the variations more pronounced which could be used to obtain rPPG. In another approach in the same work, instead of the color variations caused by the blood flow, the expansion of the blood vessels was magnified. This provided another pathway for estimating rPPG signals from the skin surface.

More recently, deep learning has been used for rPPG estimation from facial videos. In HR-CNN \cite{HRCNN}, a two-stage CNN architecture was proposed, comprising vanilla convolutions wherein the rPPG signals were estimated from the face videos and then used to predict HR. In PhysNet \cite{phys}, different spatiotemporal models based on CNNs and LSTMs were explored for rPPG estimation. In DeeprPPG \cite{deep}, a lightweight CNN architecture was used along with a novel rPPG aggregating strategy to adaptively combine rPPG signals from different skin regions. In \cite{strip}, 2D and 3D convolutions were used for the backbone architecture, followed by spatiotemporal strip pooling in the last layers to add attention to the feature maps.  

In ETA-rPPGNet \cite{eta}, a network was proposed in which a time-domain sub-network was used to reduce the redundant video information by extracting the crucial spatial features followed by a time-domain attention network to effectively predict rPPG and HR from the sub-network features. In \cite{heir}, a multi-hierarchical spatio-temporal convolutional network was proposed. In \cite{convlstm}, a two-stream architecture was proposed wherein two video inputs, the cropped face video (trunk branch) and the mask of the skin pixels (mask branch) were used. The trunk branch comprised of a combination of CNNs and Conv-LSTMs while the mask branch only had CNNs with intermediate fusion to the trunk branch through an attention mechanism for improved processing of the skin pixels. 

In \cite{deepphys}, another two-stream network was proposed where the current frame (appearance) and its normalized difference with the next frame (motion) were processed in two different CNN pathways with intermediate fusions to provide attention to the motion stream based on the appearance. This network is commonly referred to as the Convolutional Attention Network (CAN). In \cite{cdt}, a similar approach to \cite{deepphys} was proposed, but in turn replaced the standard 3D convolutions with 3D central difference convolutions (CDConv) \cite{cdc}, allowing for improved processing of the spatial and temporal information in the feature maps. In \cite{waterloo}, CAN was modified to introduce the Temporal Shift Module (TSM) \cite{tsm} for improved temporal modeling of the feature maps for rPPG estimation. In \cite{cdcrppg}, the authors used the Convolutional Block Attention Module (CBAM) \cite{cbam} to provide spatio-temporal attention in a 3D CDConv-based CNN architecture. In \cite{aaai_tfa}, the authors proposed two blocks namely the Physiological signal Feature Extraction (PFE) block and the Temporal Face Alignment (TFA) to tackle problems in rPPG estimation pertaining to changing face-camera distance and face motion.

A number of prior works have combined classical image processing techniques with CNNs. In \cite{vitasi}, phase-based video motion processing \cite{phase} was used to magnify subtle color changes and reduce the motion artifacts, followed by a CNN for remote HR estimation. In \cite{hurppg}, the video frames were first pre-processed separately using orthogonal color space projection \cite{pos} and motion normalization \cite{deepphys}, and then concatenated for processing by a CNN with different attention modules to provide spatio-temporal attention for rPPG estimation. 

In general, the deep learning approaches described above achieve effective performances. Nonetheless, given their explicit use of fully supervised training, they rely on the output labels for sustained performance.

\subsection{Self-supervised Learning}
Self-supervised learning aims to reduce the reliance of supervised learning approaches on human-annotated labels while also learning meaningful representations for enhanced performance. This training paradigm generally relies on generating pseudo-labels for pre-training neural networks prior to fine-tuning for downstream tasks. A major differentiating factor among the self-supervised approaches lies in the design of the pretext learning step. In \cite{jigsaw}, original input images were divided into several patches as puzzle pieces, and the pretext task of the network was to solve the puzzle. As a result, key visual representations and spatial consistency were learned, resulting in improved performance on the downstream task. In \cite{rotation}, the images were rotated by certain angles, and the pretext task of the network was to successfully predict these rotation angles. In \cite{inpaint}, certain regions of the image were cropped and the network was trained to fill in the regions as the pretext task. This helped the network better learn contextual information in images and perform better in subsequent downstream tasks. 

Similar to the use of \cite{bert} for self-supervision in natural language processing, masked autoencoder \cite{masked} was recently explored for self-supervised computer vision tasks. In \cite{masked}, random patches of the original image were masked and the autoencoder was trained to reconstruct the original image from the input patches. While the pre-text task is similar to \cite{inpaint}, the masked autoencoder was based on Vision Transformers (ViT) \cite{vit} instead of vanilla convolutions, allowing for the use of mask tokens \cite{bert} and positional embeddings \cite{transformer}. This helped the model learn holistic representations by encompassing the rich semantic information and be used in the downstream tasks. Furthermore, the self-supervision strategy could be scaled to high-capacity models. The paradigm of self-supervised learning has been applied to a wide variety of problems such as image classification \cite{jigsaw}, wearable-based activity recognition \cite{set1}, signal-based emotion recognition \cite{pri2, pritam22}, and more. 

Contrastive learning is a type of self-supervised learning that has gained momentum in the past few years and has shown tremendous improvement across various computer vision tasks such as object detection \cite{det}, medical image analysis \cite{med}, facial expression recognition \cite{s1,s2}, gaze estimation \cite{gaze} and others. SimCLR \cite{sim} proposed the use of different augmentations to create pseudo-samples of the original data and train the network to learn features to maximize the similarity between the augmented counterparts of the same original sample. A number of other contrastive learning approaches such as MoCo \cite{moco}, NNCLR \cite{nnclr}, have also been proposed to take advantage of different aspects of the data for learning effective representations. 

Overall, self-supervised approaches have been widely explored for a range of video-based tasks \cite{mubsurvey}. They use contrastive learning, as well as other self-supervised pre-text tasks translated directly from the image domain, as well as novel tasks designed specifically to utilize the temporal aspect of the video domain. Nevertheless, they have not been widely explored for the task of rPPG estimation.

\begin{figure}[t]
    \centering
    \includegraphics[width=1\columnwidth]{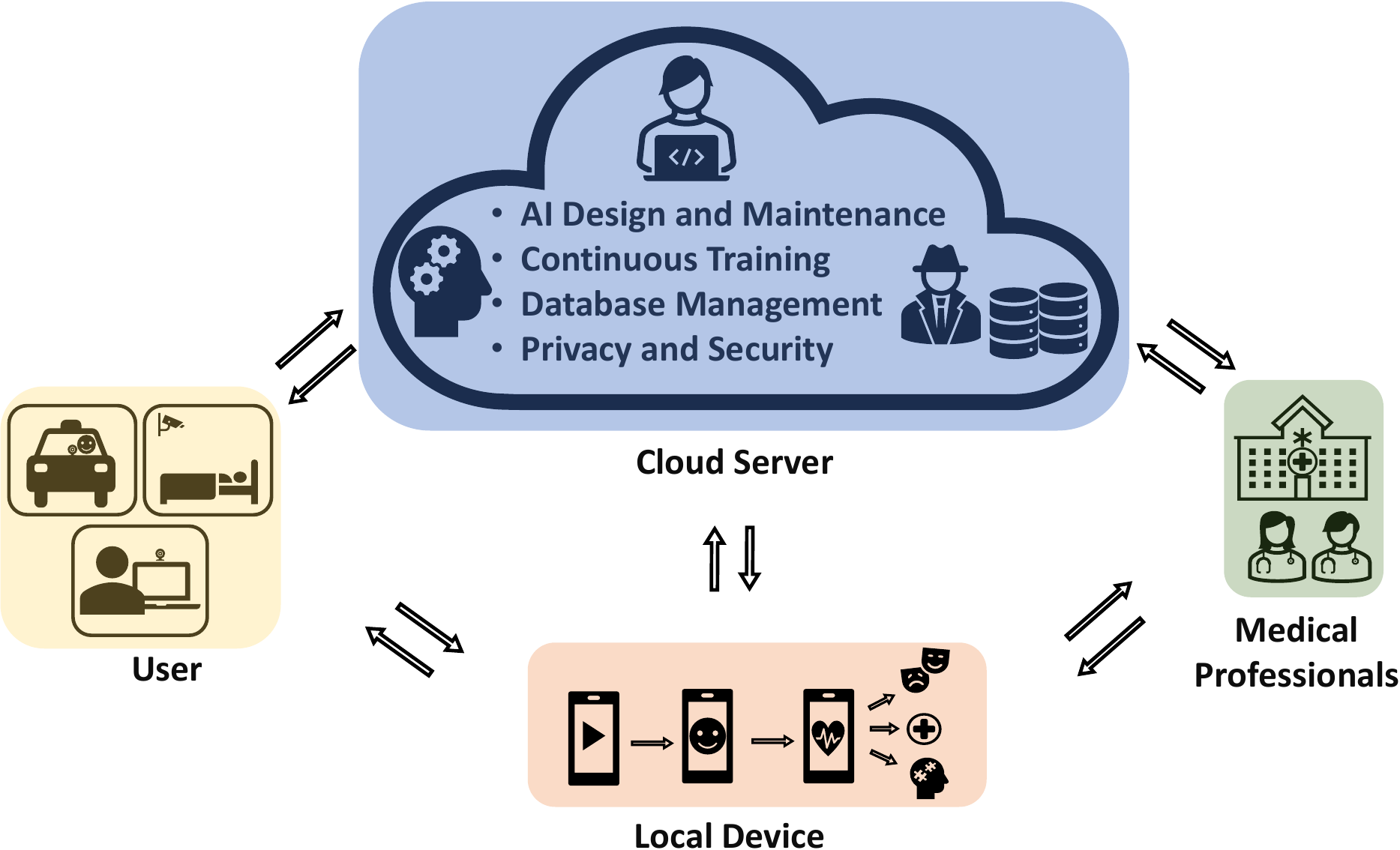}
    \caption{Depiction of a general IoT layout where our proposed method can be integrated into. The layout depicts the user, the local devices, the medical professionals, and a cloud server all working in coordination to achieve the objective of remote HR monitoring in existing smart environments.}
    \label{fig:iot}
\end{figure}

\section{System Architecture}

In this section, we discuss the integration of our work in a smart city setup. We first discuss the various instances wherein our work can be integrated into different smart environments, following which we discuss the different components used by our solution in a typical IoT framework.

\subsection{Smart-City Contextualization}
A smart city constitutes a variety of smart environments such as smart hospitals, smart homes, smart workplaces, smart vehicles, and others. While each of these environments processes different information for different purposes, on a rudimentary level, they comprise similar components namely data acquisition devices, communication channels, and cloud servers. Moreover, these environments generally work towards the common goal of utilizing advanced technologies to add value and convenience to the lives of the citizens of a smart city and enhance their quality of life \cite{city}. 

Our work falls primarily in line with remote health monitoring which has become an integral part of various smart environments. With the adoption of the remote health monitoring paradigm, medical professionals can monitor the vitals and other physical symptoms of a patient without having to be in the same place as the patient and provide them with adequate consultation. 

In smart environments such as smart homes and smart workplaces, remote monitoring of vitals helps to monitor health without having to leave the premises or commute to medical institutions \cite{smart_home, smart_work}. This benefits especially those who lack mobility such as the senior and physically challenged people \cite{smart_old, smart_challenge}. In smart vehicles too, the vitals of the user can be tracked and appropriate feedback could be provided on the go \cite{smart_vehicle}. These varied use cases discussed, further the impact and importance of our work. 

In Figure \ref{fig:iot}, we illustrate the integration of our work into an existing IoT layout. The layout comprises the user in various surroundings with access to a camera, local devices to run inference algorithms, medical professionals to provide consultations, and lastly a centralized cloud server for continuous training and maintenance of data and/or algorithms. The user can run a remote vitals checkup in the layout as long as they have access to a camera for capturing the video of their face. Next, our proposed algorithm would be run on any available local device for inferring the rPPG signal from the face video. After estimating the rPPG, further insights such as vitals, emotions (for mood and mental health management), and others can be derived from it and be made available to the user and appropriate medical consultation can be provided if needed. All these processes would take place in conjunction with the cloud server. For capturing face video, a good camera which is common in smart environments will suffice. For computing on local devices, deep learning algorithms have been known to be deployed on a variety of devices besides computers, e.g., mobile devices and micro-controllers \cite{mobile,ras_pi}, which makes the integration of our work into any existing IoT system seamless, allowing for it to be used across multiple environments.   

Lastly, since our proposed method involves the training of intelligent algorithms, the contextualization can also be broadened to base upon the cognitive IoT (CIoT) \cite{ciot} paradigm. The CIoT paradigm is an upgrade over the IoT paradigm as it incorporates intelligence into the existing IoT components, thereby adding another layer of `smartness' in the smart environment. Despite that, for simplicity as done in \cite{arthur}, we too adopt the general IoT paradigm for the contextualization of our work in the next section.

\subsection{IoT Architecture}
\begin{figure}[t]
    \centering
    \includegraphics[width=1\columnwidth]{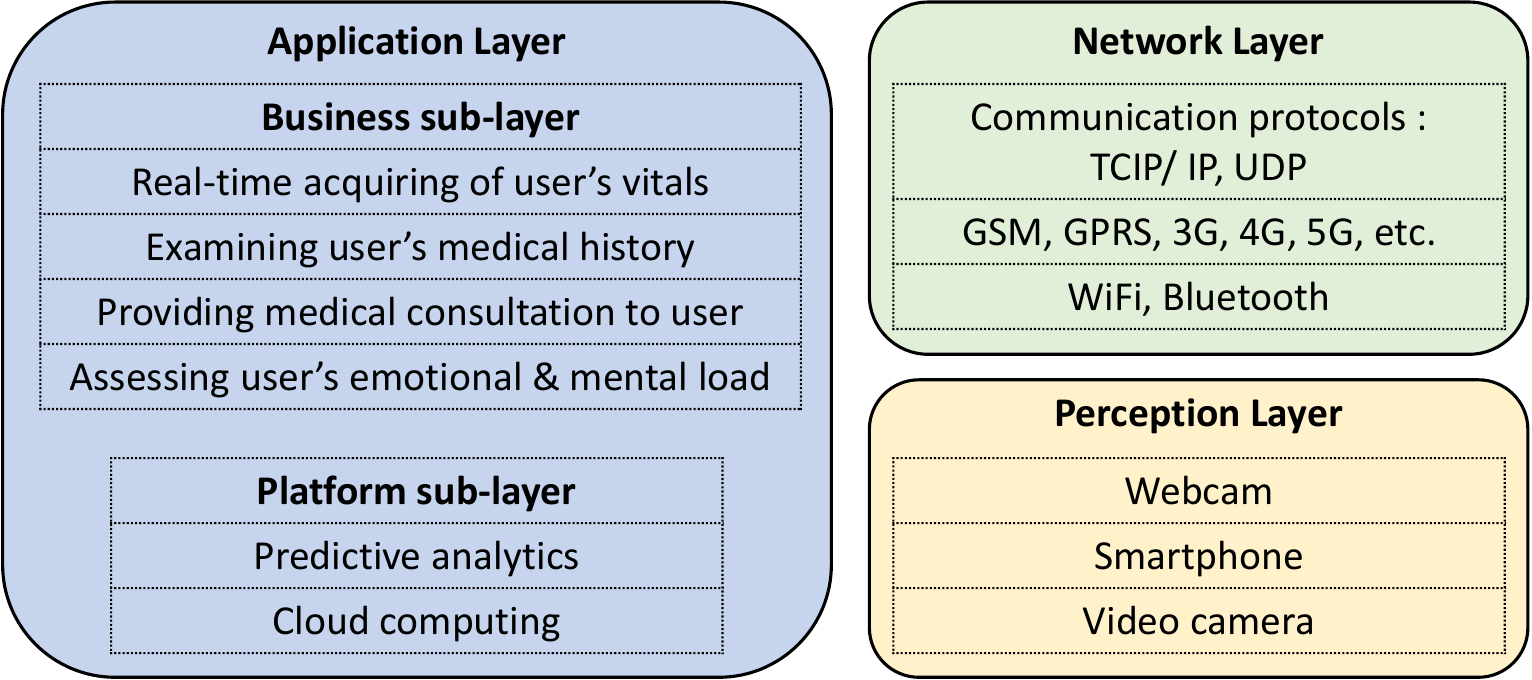}
    \caption{A typical 3-layered IoT architecture in the context of our remote HR solution. The architecture comprises of the Application Layer, the Network Layer, the Perception Layer, and their constituents.}
    \label{fig:layer}
\end{figure}

IoT architectures are often modeled using three layers of abstraction, namely the perception layer, the network layer, and the application layer \cite{iotarch}. In the following, we describe our work on remote HR estimation in the context of an IoT architecture for smart environments with consideration of these three layers. An overview of the architecture is shown in Figure \ref{fig:layer}.

\noindent \textbf{Perception Layer.} This layer serves as the information source of an IoT system and involves data acquisition devices. In our work, this layer comprises the various cameras present in the smart environments. Examples include cameras in smartphones, webcams on computers, driver/passenger-facing cameras in vehicles, video cameras in smart homes, and others. Any of these can be used to record a video of the face to be used subsequently for inference. 

\noindent \textbf{Network Layer.} This layer is responsible for conveying the information from the perception layer to the application layer. This layer is essentially based on the existing Internet and mobile telecommunication infrastructure. Some examples of the components include general packet radio service (GPRS), fourth/fifth-generation (4/5G) communication, WiFi, and others that provide wireless and long-distance communication. For our proposed solution, the Internet can be accessed through smart computing devices such as smartphones, computers, smart camera systems, and others. Also, Bluetooth can be used by smart devices for short-distance communication.

\noindent \textbf{Application Layer.} This layer processes the information received from the perception layer into useful applications. It can be further divided into platform and business sub-layers. The platform sub-layer comprises various algorithms and protocols designed, run, and updated on the cloud to ensure the smooth functioning of the entire architecture. Mainly four functions are carried out at the platform sub-layer:
    (1) AI Design and Maintenance, 
    (2) Continuous Training, 
    (3) Database Management, and
    (4) Privacy and Security.

Our proposed method would be a part of this sub-layer which would be used for inference of heart rate remotely and without the need for physical sensors to come in contact with the user and also undergo subsequent training on new data. Other models used to derive vitals and other information from the rPPG signals too are a part of this layer. Besides the engineering team, medical professionals would also contribute to this sub-layer by providing field expertise to better guide the designing of the algorithms. The business sub-layer uses the final extracted information to meet the goals and requirements of the stakeholders. The user as well as the medical professionals could be notified of the user’s vitals if/when required and also be alerted appropriately in the case of any anomalies to ensure prompt diagnosis of any symptoms. Besides remote health monitoring, other applications such as stress assessment \cite{work_stress} at workplaces or emotion recognition \cite{driver_emotion} while driving can be conceptualized. Since this layer is very crucial to the working of the entire architecture, it would be paid extra attention to secure it safely and to ensure all-time connectivity. 

A key aspect of our overall envisioned architecture for remote PPG monitoring in smart environments is the notion of ubiquity, while also maintaining and protecting user privacy. As a result, as we show in the following subsections, we design our deep learning solution with this notion in mind, where we utilize specific regions of the face (cheeks and forehead) for rPPG estimation. This approach would allow for the user's full facial image to not be entered into the pipeline, therefore, allowing for better privacy preservation. Additionally, in the IoT design, special attention should be given to utilizing privacy-preserving approaches such as federated learning \cite{fed}, data perturbation methods \cite{divij_smc}, and secure software and cloud practices \cite{sherman_security} to ensure user security and privacy.

\section{Proposed Method} 

\subsection{Solution Architecture}
\begin{figure*}[t]
    \centering
    \includegraphics[width=2\columnwidth]{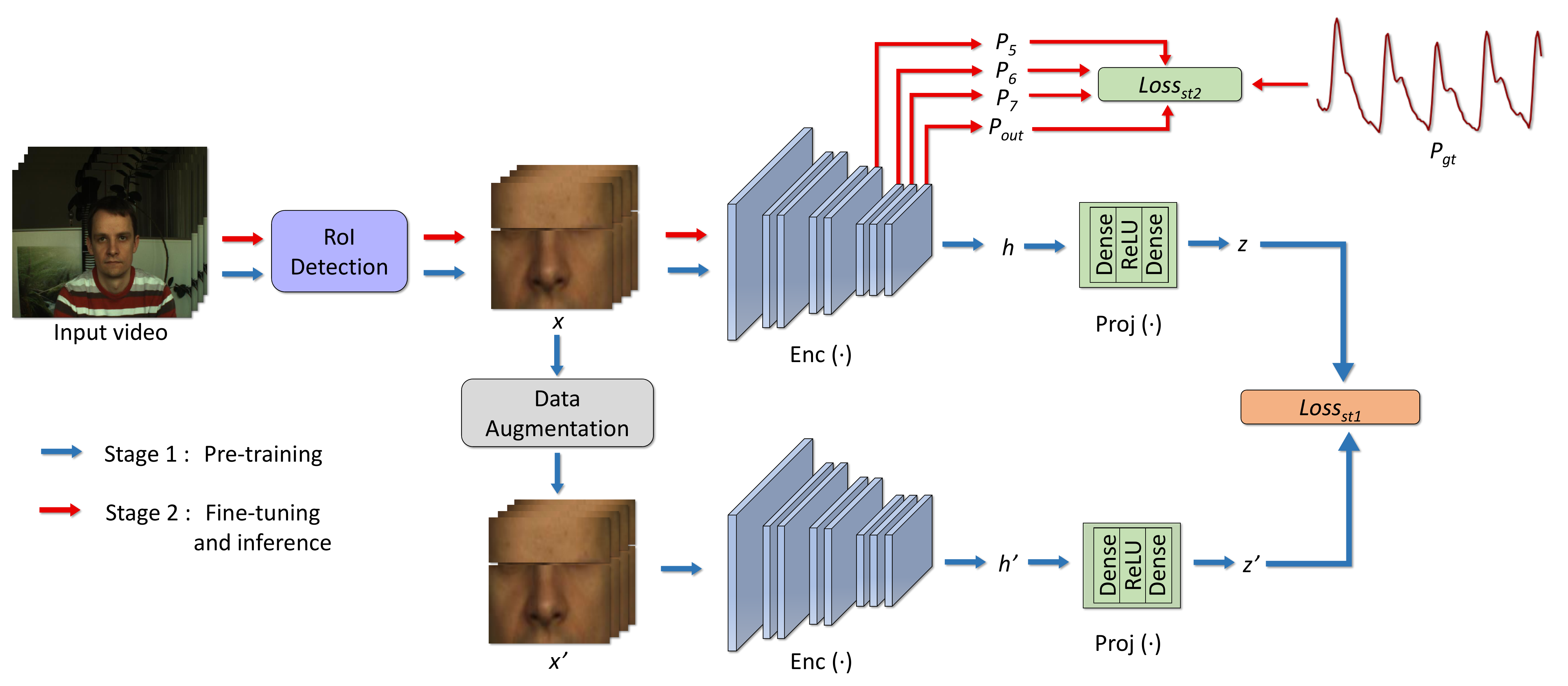}
    \caption{The overall layout of the proposed two-stage approach. The first stage, denoted using the blue arrows, uses unlabeled data (video clips) for the contrastive pre-training of the encoder while the second stage, shown with red arrows, involves fine-tuning of the encoder using labeled data (video clip and PPG) for rPPG estimation. The RoI detection module is described in Section \ref{roidet}. Enc(.) denotes the encoder, while Proj(.) denotes the projection head consisting of a dense layer, a ReLU activation layer, and another dense layer. All the modules including data augmentation, encoders, and projection heads are described in  Section \ref{simclr}. Finally, $Loss_{st1}$ and $Loss_{st2}$ denote the stage 1 and stage 2 losses described in Eqs. \ref{contr_eq} and \ref{st2lss} respectively.}
    \label{fig:arch}
\end{figure*}

Our method consists of two separate main stages: (1) self-supervised contrastive pre-training, and (2) supervised fine-tuning. First, we take a raw input video clip and detect regions of interest (RoI), namely the forehead and cheeks. This is done for two main reasons, first to enable more robust rPPG estimation, and second to allow for better user privacy protection. Next, subsequent to the detection of the RoI, we enter the first stage of our method where the RoI clip is processed by a data augmentation module to generate an augmented RoI clip. After this, the RoI and its augmented counterpart are passed through an encoder and subsequently, the projection head to generate lower-level feature embeddings. This is done for all the input RoI clips. The contrastive loss is then used to learn strong representations by maximizing the similarity between embeddings belonging to the same RoI clip while minimizing the similarity between embeddings from separate RoI clips. Subsequently, in stage 2, we use the encoder from stage 1 and fine-tune it using the RoI clip as the input and the corresponding PPG signal as the output via smooth L1 loss. The architecture of our solution is illustrated in Figure \ref{fig:arch}. Through the following subsections, we describe each component of our solution mentioned above, in detail.

\subsection{Pre-processing}\label{roidet}

Since observable changes in blood flow, and thus rPPG, are stronger around the forehead and the cheek regions \cite{ppgroi, ppgroi2}, we detect and crop these regions as our RoI, using the Dlib-face Detector \cite{dlib} (see Figure \ref{fig:arch}). Next, we concatenate these regions for each frame and resize the outcome to 64$\times$64 pixels. For each video, we use a sliding window with a length of 128 frames and a stride of 8 frames to obtain several smaller clips. Similarly, we segment the ground-truth PPG signals such that the time-synchronicity between each video clip and the corresponding PPG segment is maintained (this will be used in the supervised fine-tuning stage). This cropping of the RoI also provides a layer of security as it does not use the regions with high levels of discrimination in facial recognition such as the periocular region, the lips, and others \cite{bioeye,biolip,facer} while using the regions with relatively lower levels of discrimination namely the forehead and cheek \cite{imperf}.

\vspace{1em}
\subsection{Stage 1: Self-supervised Pre-training}  \label{simclr}
As mentioned earlier, our self-supervised pre-training step consists of a data augmentation module, an encoder, and a projection head. Here we describe each component in detail.

\noindent \textbf{Data Augmentation.} This module applies a set of augmentations to the input RoI clip $x$ with $M$ frames, $x_1, x_2,... x_M$, to generate $x'$ which also consists of $M$ frames. In the proposed method, we use two categories of augmentations: (\textit{i}) spatial, and (\textit{ii}) temporal. In terms of spatial augmentations, we employ the following:
\begin{itemize}

    \item \textit{Rotation}, where all the frames are rotated by the same angle $\theta \in \mathbb{N}$, where $\theta$ is chosen randomly from $\{1, 2 ,..., 360\}$;
    
    \item \textit{Crop}, where for a frame with height $H$ and width $W$, we randomly select a cropping scale $\gamma \in \mathbb{R}$ from $[0.25, 0.75]$, and choose the cropping window anchor with coordinates $i \in \mathbb{N}$ and $j \in \mathbb{N}$. The anchor coordinates are chosen randomly from $\{0, 1,... , W - \gamma W\}$ and $\{0, 1,... , H - \gamma H\}$ respectively. Accordingly, the crop is performed between $(i,j)$ and $(i+\gamma W, j+\gamma H)$, followed by resizing of the output to $W \times H$;
    
    \item \textit{Flip}, where every frame is flipped along the vertical axis. Mathematically, for pixel value with coordinate $(i,j)$ in frame $x_m'$ from $x'$ and corresponding frame $x_m$ from $x$, we have $x'_m(i,j) = x_m(W-i,j)$.
\end{itemize} 

As mentioned, we also perform temporal augmentations, as follows:
\begin{itemize}
    \item \textit{Shuffle}, where frames $x_1, x_2,... x_M$ are shuffled randomly to obtain $x'$ with a different order of frames $x_{1'}, x_{2'}, x_{3'}, ... x_{M'}$;
    
    \item \textit{Reorder}, where a random index $r$ is selected to cut the video into two clips $x_a = x_{1}, x_{2}, ..., x_{r-2}, x_{r-1}$; $x_b = x_{r}, x_{r+1}, x_{r+2}, ... x_{M-1}, x_{M}$. $x'$ is then synthesized as $x' = [x_b, x_a]$;

    \item \textit{Reverse}, where the order of frames is reversed to obtain $x' = [x_{M}, x_{M-1}, ...,$ $x_{2}, x_{1}]$.
\end{itemize} 
 
In our experiments, input RoI clip $x$ and its augmented counterpart $x'$ make up a positive pair between which the similarity is maximized with contrastive learning. Alternatively, for two different input clips $x_1$ and $x_2$, where $x_1 \neq x_2$, the samples $(x_1,x_2)$, $(x'_1,x_2)$, $(x_1,x'_2)$, and $(x'_1,x'_2)$ constitute the negative pairs, between which the similarity is minimized. This is done for all the input clips in the batch and hence, all the input video clips of the training dataset in an epoch.

\begin{table}
\centering
\caption{Architectural details of the encoder used in our proposed method.}
\small
\begin{tabular}{c c c} 
\hline
\textbf{Layer~Name} & \textbf{Kernel~Size}      & \textbf{Output Size}        \\ \hline\hline
Input               & -             & 128$\times$64$\times$64$\times$3      \\ \hline
Conv1               & 16, [1,5,5]            & 128$\times$62$\times$62$\times$16     \\ \hline
AvgPool              & {[}1,2,2]               & 128$\times$31$\times$31$\times$16      \\ \hline
ConvBlock1          & \begin{tabular}[c]{@{}c@{}}32, [3,3,3]\\32, [3,3,3]\end{tabular}          & \begin{tabular}[c]{@{}c@{}} 128$\times$31$\times$31$\times$32\\ 128$\times$31$\times$31$\times$32\end{tabular} \\ \hline
AvgPool               & {[}1,2,2]              & 128$\times$15$\times$15$\times$32      \\  \hline
ConvBlock2          & \begin{tabular}[c]{@{}c@{}}64, [3,3,3]\\64, [3,3,3]\end{tabular}  & \begin{tabular}[c]{@{}c@{}}128$\times$15$\times$15$\times$64\\128$\times$15$\times$15$\times$64\end{tabular}    \\  \hline
AvgPool               & {[}1,2,2]               & 128$\times$7$\times$7$\times$64      \\ \hline
ConvBlock3          & \begin{tabular}[c]{@{}c@{}}64, [3,3,3]\\64, [3,3,3]\\64, [3,3,3]\end{tabular}  & \begin{tabular}[c]{@{}c@{}} 128$\times$7$\times$7$\times$64\\ 128$\times$7$\times$7$\times$64\\  128$\times$7$\times$7$\times$64\end{tabular}  \\ 
\hline
Global AvgPool              & 1, [1, 7, 7]            & 128$\times$1$\times$1$\times$64        \\ \hline
Squeeze              &  -           & 128$\times$  64  \\
\hline
Output              & 1, [1]             & 128$\times$1        \\
\hline
\end{tabular}
\label{config}
\end{table}

\noindent \textbf{Encoder.} 
We use a 3D CNN architecture as our encoder. The initial input is passed through a 1$\times$5$\times$5 kernel that tends to extract information from each video frame. Next, our model performs 3D convolutions with kernel 3$\times$3$\times$3 on the resulting embeddings. The detailed architecture is given in Table \ref{config}. Each convolution operation is followed by a ReLU activation and batch normalization. Mathematically, for any input $x$, $h = Enc(x)$, where $Enc(\cdot)$ is the encoder and $h$ is the intermediate embedding of $x$.

\noindent \textbf{Projection Head.} 
Following the encoder, we use a projection head to map the obtained embedding onto a lower-dimensional space. To this end, we use a 2-layer dense neural network with 64 and 16 neurons to generate the low-dimensional embedding of the output of the encoder. The final embedding, $z$ is given by $z = Proj(h)$, where $Proj(\cdot)$ is the projection head. 

\noindent \textbf{Contrastive Loss.}
We use the contrastive loss presented in \cite{sim} for the pre-training stage of our model. This loss helps in learning representations that maximize the similarity between positive pairs while minimizing the similarity between negative samples. For any pair of clips $(x_m,x_n)$ with corresponding projections $(z_m,z_n)$, the cosine similarity is given by:
\begin{equation}
cosine(z_m, z_n) = \frac{z_m^T z_n}{||z_m||.||z_n||} ,
\end{equation}
where $||.||$ denotes the L2 norm (magnitude) of the projections. Subsequently, the loss function is given as:
\begin{equation}
\label{contr_eq}
Loss_{st1}(z_m,z_n) =  - log \frac
{exp(cosine(z_m,z_n)/\tau)}
{\sum_{k=1}^{2N} \mathbbm{1}_{[k \neq m]} exp(cosine(z_m,z_k)/\tau)} ,
\end{equation} 
where $\mathbbm{1}_{[k \neq m]} \in \{0,1\}$ is the indicator function that outputs 0 iff $k=m$ and 1 otherwise, $\tau$ is the temperature hyper-parameter, and $2N$ is the total number of samples resulting from augmenting the original $N$ samples.

\subsection{Stage 2: Supervised Fine-tuning} 
For the second stage of the proposed method, we discard the projection head $Proj(.)$ and the data augmentation module from the previous stage and only use the encoder. We fine-tune the entire encoder by using the RoI clip as the input and the PPG signal as the output. Furthermore, instead of using the output of only the last layer of the encoder for training, we use the output embeddings of the final four convolutional layers. This enables more effective representations to be learned throughout different parts of the encoder. 

\noindent \textbf{Smooth L1 Loss.}
We use the smooth L1 loss \cite{fastrcnn} for the second stage of training. This loss is a combination of both L1 and the L2 losses and allows for switching between the two depending upon the difference between the amplitude values of the rPPG signal $P_{out}$, and the ground-truth PPG signal $P_{gt}$. This loss is given by:
\begin{equation}
\mathcal{L}(P_{out}, P_{gt}) =
    \begin{cases}
      \frac{1}{2} \frac{(P_{out}-P_{gt})^2}{\beta}, ~~ |P_{out} - P_{gt}| < \beta \\
      |P_{out} - P_{gt}| - \frac{1}{2}*	\beta, ~~ otherwise .
    \end{cases} 
\end{equation}
where $\beta$ is a threshold hyperparameter. Here, when the absolute difference between the estimated and ground-truth signals is smaller than $\beta$, the loss uses the L2 loss, otherwise, it uses L1. The L2 loss is quite sensitive to large errors due to its square operation. Thus, to obtain a smooth output, i.e., more effective training, the loss shifts to L1 for signals with larger differences. As mentioned earlier, we apply this loss to the output embeddings of the final four convolutional layers as opposed to only the final layer, which enables more effective representations to be learned throughout different parts of our network. Accordingly, we calculate the final loss for stage 2 by:
\begin{equation}
\label{st2lss}
Loss_{st2} = \mathcal{L}(P_{out}, P_{gt}) + \alpha \sum^{\lambda2}_{i=\lambda1} \mathcal{L}(P_i, P_{gt}),
\end{equation}
where $P_i$ is the output of the $i^\text{th}$ convolutional layer. We empirically set $\lambda1 = 5$ and $\lambda2 = 7$ as the outputs from the 5$^\text{th}$, 6$^\text{th}$, and 7$^\text{th}$ layers provided valuable auxiliary information, in addition to the final layer $P_{out}$. Finally, we empirically set the weight of the auxiliary loss terms, $\alpha$, to 0.5.

\noindent \textbf{HR Calculation.}
After obtaining the estimated rPPG signals at runtime, similar to \cite{HRCNN,deep} we calculate the HR by measuring the largest peak obtained from the Welch power spectrum of the signal. 

\subsection{Implementation}
The batch sizes for stage 1 and stage 2 of the training are set to 16 and 8, where we train the model for 50 and 10 epochs, respectively. The learning rates are set to 1e-4 and 2e-4 for stages 1 and 2, respectively, with Adam used as the optimizer for both. The value of $\beta$ is taken as 1 for one of the datasets (COHFACE) and 0.3 for the other (PURE). All the hyper-parameters were chosen empirically to maximize performance. All the codes are written in PyTorch and run on an NVIDIA GTX 2080 Ti GPU.

\section{Experiment Setup}
In this section, we first describe the datasets used in our study. This is followed by the evaluation scheme and the metrics used for comparison of our solution to prior work. And finally, we discuss in detail the variations of our method which we use to validate our design choices. 
\subsection{Datasets}
\label{datasets}
We use two publicly available datasets, COHFACE \cite{cohface} and PURE \cite{pure}, for our experiments. Following is a description of each dataset.
\begin{itemize}
    \item \textbf{COHFACE \cite{cohface}:} This dataset comprises 160 facial videos and their corresponding PPG. There are a total of 40 subjects (28 males, and 12 females) and each subject contributes 4 videos. The videos have been recorded under two illumination settings (natural lighting and studio lighting). In the natural lighting setting, the face of the subject is unevenly illuminated by the light coming from the window blinds next to the subject. In the case of studio lighting, the face is evenly illuminated by the ceiling light and a 400W halogen spotlight. The videos have been recorded with a Logitech HD Webcam C525 at 20 fps while the blood volume pulse signals have been recorded using a contact TTL SA9308M sensor at a 256 Hz sampling rate. The videos were compressed in MPEG-4 format with a resolution of 640$\times$480 pixels.  

    \item \textbf{PURE \cite{pure}:} This dataset comprises 60 facial videos and their corresponding PPG. There are a total of 10 subjects (8 males, and 2 females) and each subject contributes 6 videos performing 6 different movements. The movements are steady sitting, talking, small face rotation, medium face rotation, slow face translation, and fast face translation. The videos have been recorded with an Eco274CVGE camera at 30 fps while the PPG signals have been recorded using a finger pulse oximeter Pulox CMS50E at a 60 Hz sampling rate. The videos have been stored with lossless compression in PNG format with a resolution of 640$\times$480 pixels. 
    
\end{itemize}

\begin{figure}
\centering
  \includegraphics[width=1\linewidth]{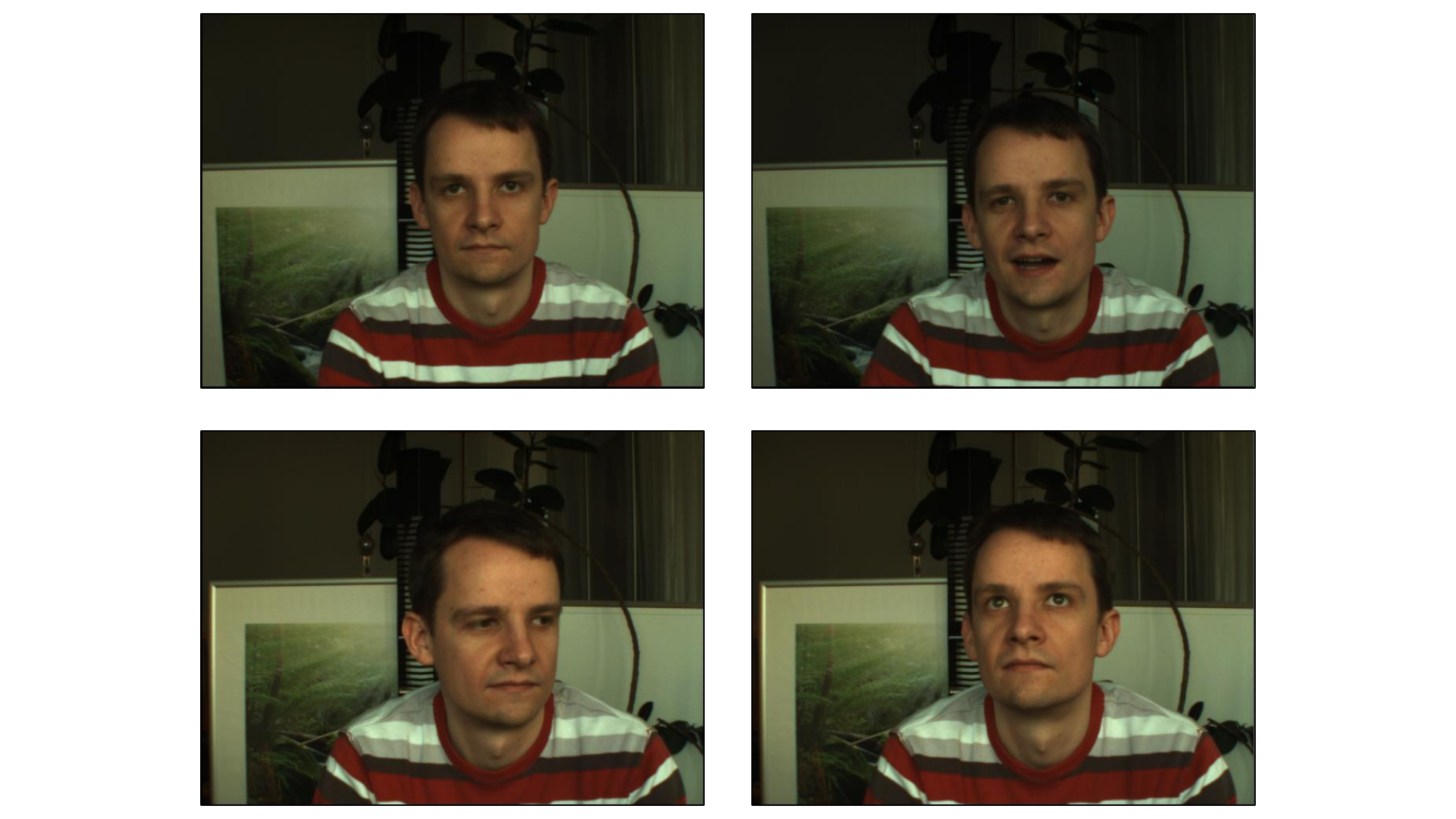}
\caption{Sample frames showing the varying conditions such as sitting, talking, face translation, and rotation in the PURE dataset.}
\label{frames}
\end{figure}

\subsection{Evaluation Scheme and Metrics}
For COHFACE, we use the subject split that has been designated and provided by the original authors of the dataset \cite{cohface}. Specifically, the data from 24 subjects is used for training our model, while the data from the remaining 16 subjects is used for testing. For PURE, we use a 6-4 subject train-test split as commonly used in prior works such as \cite{HRCNN, deep, cdt}.

To evaluate our model, the HR values obtained from the individual RoI clips of the same test video are averaged to generate one HR for each test video. These averaged HR values are then compared with the actual average HR calculated from the ground-truth PPG signals. The metrics we use for evaluation are, mean absolute error (MAE) and root mean square error (RMSE), both in beats per minute (bpm), along with correlation (\textit{R}) of the predicted HR ${HR_{pred}}$ and ground-truth HR ${HR_{gt}}$. For $N$ test videos, the metrics are calculated as: 
\begin{equation}
MAE = \frac{1}{N} \sum^{N}_{i=1} |HR_{pred}(i) -  HR_{gt}(i)|,
\end{equation}
\begin{equation}
RMSE = \sqrt{\frac{1}{N} \sum^{N}_{i=1}(HR_{pred}(i) -  HR_{gt}(i))^2},
\end{equation}
and 
\begin{equation}
\small
R = \frac{\sum^{N}_{i=1}(HR_{pred}(i) -  \overline{HR_{pred}}) (HR_{gt}(i) - \overline{HR_{gt}})}{\sqrt{\sum^{N}_{i=1}(HR_{pred}(i) -  \overline{HR_{pred}})^2\sum^{N}_{i=1}(HR_{gt}(i) -  \overline{HR_{gt}})^2}},
\end{equation}
where $\overline{HR_{pred}}$ and $\overline{HR_{gt}}$ represent the mean of the estimated and ground-truth average HR of the test videos respectively.

\subsection{Comparisons}
Here we briefly describe the other methods with which we compare our proposed solution. 

\noindent \textbf{Prior Works.}
We compare our work with several previous works discussed in Section \ref{related}. These works include \cite{pos, chrom} which use classical image and signal processing approaches, as well as a large number of deep learning approaches. The deep learning approaches include methods with vanilla architectures such as \cite{HRCNN,deep}, two-stream approaches such as \cite{deepphys,convlstm}, varied attention mechanisms \cite{eta,strip}, methods combining classical approaches with deep learning such as \cite{hurppg,vitasi}, and others.

\begin{table}
\centering
\caption{Architectural details of the (2+1)D encoder (Encoder B) used in our experiments.}
\small
\begin{tabular}{c c c} 
\hline
\textbf{Layer~Name} & \textbf{Kernel~Size}      & \textbf{Output Size}        \\ \hline\hline
Input               & -             & 128$\times$64$\times$64$\times$3      \\ \hline
Conv1               & 16, [1,5,5]            & 128$\times$62$\times$62$\times$16     \\ \hline
AvgPool              & {[}1,2,2]               & 128$\times$31$\times$31$\times$16      \\ \hline
ConvBlock1          & \begin{tabular}[c]{@{}c@{}}57, [1,3,3]\\32, [3,1,1] \\ 72, [1,3,3]\\32, [3,1,1]\end{tabular}    & \begin{tabular}[c]{@{}c@{}} 128$\times$31$\times$31$\times$57\\ 128$\times$31$\times$31$\times$32 \\ 128$\times$31$\times$31$\times$72\\ 128$\times$31$\times$31$\times$32\end{tabular} \\ \hline
AvgPool               & {[}1,2,2]              & 128$\times$15$\times$15$\times$32      \\  \hline
ConvBlock2          & \begin{tabular}[c]{@{}c@{}}115, [1,3,3]\\64, [3,1,1] \\ 144, [1,3,3]\\64, [3,1,1] \end{tabular}    & \begin{tabular}[c]{@{}c@{}}128$\times$15$\times$15$\times$115\\128$\times$15$\times$15$\times$64 \\ 128$\times$15$\times$15$\times$144\\128$\times$15$\times$15$\times$64\end{tabular}    \\  \hline
AvgPool               & {[}1,2,2]               & 128$\times$7$\times$7$\times$64  \\ \hline
ConvBlock3          & \begin{tabular}[c]{@{}c@{}}144, [1,3,3]\\64, [3,1,1] \end{tabular} \}$\times$3  & \begin{tabular}[c]{@{}c@{}} 128$\times$7$\times$7$\times$144\\ 128$\times$7$\times$7$\times$64\end{tabular} \}$\times$3  \\ 
\hline
Global AvgPool              & 1, [1, 7, 7]            & 128$\times$1$\times$1$\times$64        \\ \hline
Squeeze              &  -           & 128$\times$  64  \\
\hline
Output              & 1, [1]             & 128$\times$1        \\
\hline
\end{tabular}
\label{tab:21D}
\end{table}

\begin{figure}[t]
    \centering
    \includegraphics[width=1\linewidth]{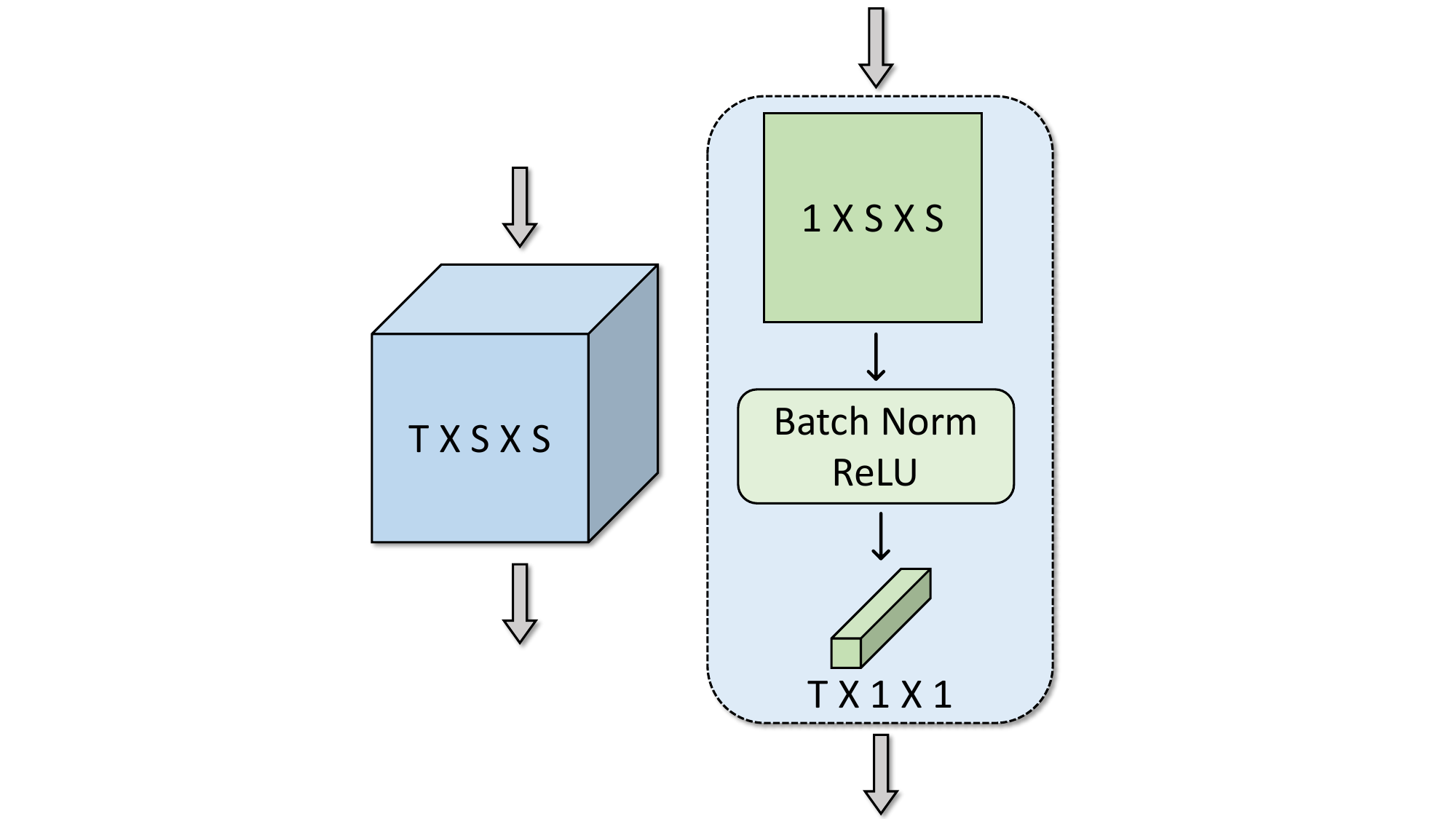}
    \caption{Illustration of the 3D and (2+1)D convolutions. T stands for the temporal filter size, while S stands for the spatial filter size. The 3D convolution processes the temporal and spatial information uniformly together while the (2+1)D convolution processes the temporal and spatial information separately and sequentially.}
    \label{fig:213d}
\end{figure}

\noindent \textbf{Video Representation Learning.} 3D convolutions are a common convolutional approach for processing 3D data such as videos. The 3D convolution does not distinguish among the dimensions of the data and treats the different dimensions equally. However, there are convolutional units such as \cite{21d, cdc} and others that tend to break down the processing of spatio-temporal data across the dimensions to better process the information. Of these, the (2+1)D convolution is widely used in video-based supervised as well as self-supervised learning \cite{s1,21dexamp,deep}. In a (2+1)D convolution, the 3D convolution is decomposed into a combination of spatial (2D) and temporal (1D) convolutions. The 2D convolution first extracts the spatial features from the input, after which the 1D convolution extracts the temporal features from these intermediate embeddings to complete the spatio-temporal processing. The separate processing is appropriate for rPPG estimation since there is less spatial variation in the facial videos as compared to the temporal one. There are also additional non-linearities (batch-normalization and ReLU) introduced in the intermediate step which helps in learning better representations. While our main solution uses 3D convolutions, for comparison purposes, we follow \cite{21d} to implement the (2+1)D approach. We use kernel sizes of 1$\times$3$\times$3 and 3$\times$1$\times$1 for the spatial and temporal convolution respectively. A detailed layout of the (2+1)D version of the encoder is presented in Table \ref{tab:21D} and depicted in Figure \ref{fig:213d}.

\noindent \textbf{Negative Pairs for Self-supervised Pre-training.}
We also study the effect of using negative pairs in the self-supervised pre-training stage of our proposed method. While SimCLR uses both positive and negative pairs to train the self-supervised learning algorithm, there are other self-supervised techniques that do not use negative pairs. One such self-supervised paradigm is SimSiam \cite{simsiam}, which we adopt for comparison purposes.

\begin{figure}[t]
    \centering
    \includegraphics[width=1\linewidth]{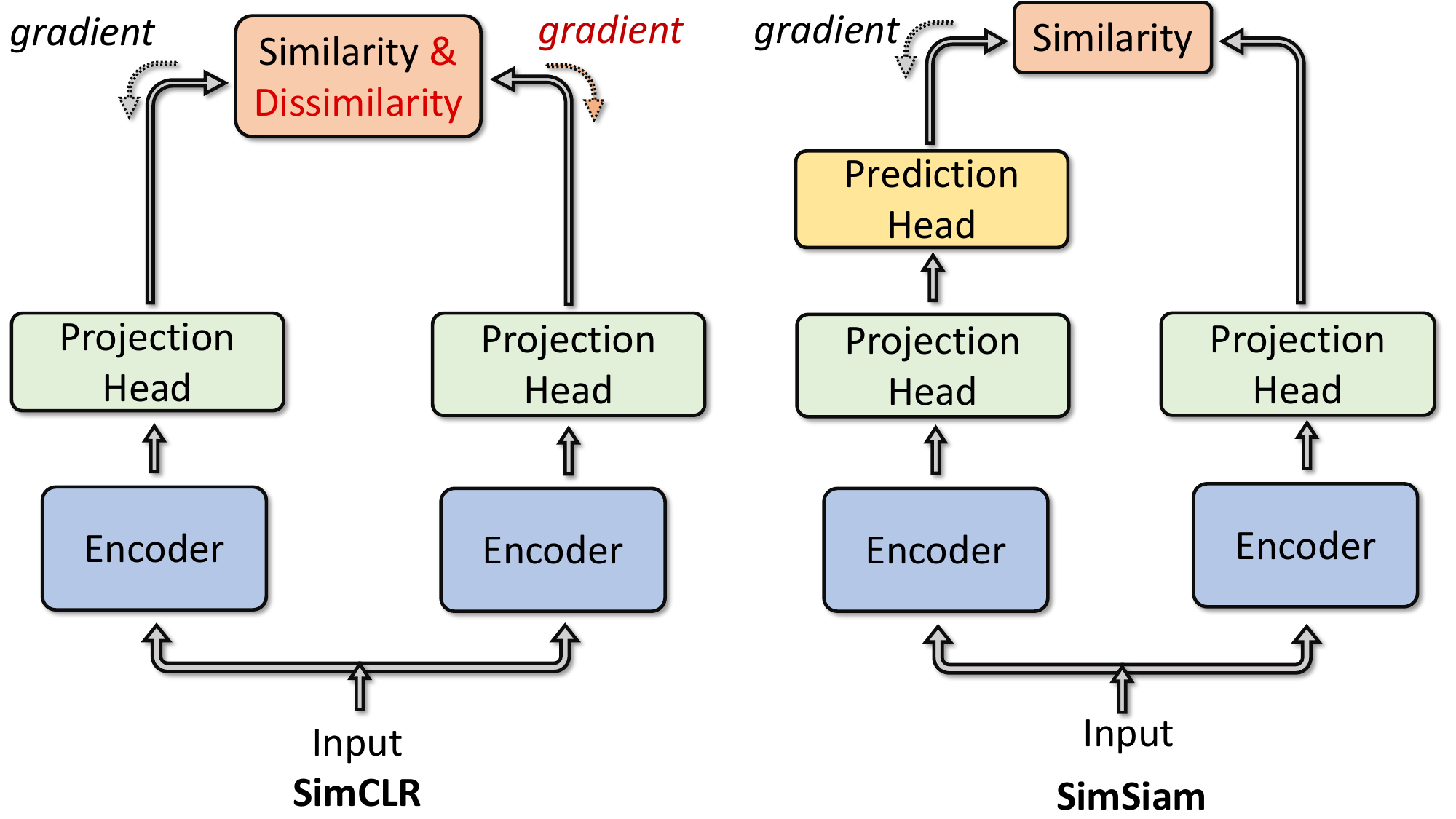} 
    \caption{An overview of the self-supervised learning strategies used in this study. On the left is SimCLR which maximizes the similarity between the positive pairs and minimizes the similarity between the negative pairs, while on the right is SimSiam which only maximizes the similarity between the positive pairs.}
    \label{fig:simm}
\end{figure}

The framework used in SimSiam is similar to SimCLR, but with the inclusion of another dense network, the prediction head or the predictor. Similar to the projection head discussed in Section \ref{simclr}, the prediction head is used to map the lower level embedding $z$ to another embedding space such that $p = Pred(z)$, where $p$ is the prediction embedding and $Pred(.)$ is the prediction head. SimSiam trains the model such that the predictor learns to predict the representation of one view of the input such that it is similar to the projection of another. Through this, the essential features of the input are learned by the model. SimSiam uses the predictor in only one branch of the model while applying a stop-gradient to the other. In this manner, the projection of one view of the input is constant with respect to the prediction of the other. The objective of SimSiam is to minimize the negative cosine similarity $D$ between the prediction embedding $p$ and the projection embedding $z$ given by: 
\begin{equation}
D(p, z) = - \frac{p^T.z}{||p||.||z||} .
\end{equation}
To ensure that both views of the input are processed by both branches of the framework, the loss function is symmetrized. Therefore, for a positive input pair of clips $(x_m,x_n)$, the loss is given as:
\begin{equation}
\small
Loss_{sim}(m, n) = \frac{1}{2} D(p_m, stopgrad(z_n)) + \frac{1}{2} D(p_n, stopgrad(z_m)),
\end{equation}
where $(p_m,p_n)$ and $(z_m,z_n)$ are the predictions and projections of $(x_m,x_n)$ and $stopgrad(.)$ is the stop-gradient function discussed above. An overview of the key differences among the self-supervised learning approaches is presented in Figure \ref{fig:simm}. For SimSiam, we use a 3-layer dense neural network with 64, 32, and 32 neurons as the projection head and a 2-layer dense neural network with 8 and 32 neurons as the prediction head. After pre-training, we followed the same procedure as our proposed method for fine-tuning the encoder for the rPPG estimation.

\begin{table*}[!t]
\centering
\caption{Comparison of our proposed method with prior works on COHFACE, PURE, and PURE (MPEG-4) datasets.}
\small
\setlength
\tabcolsep{6pt}
\begin{tabular}{l|c c c|c c c|c c c}
\hline
 & \multicolumn{3}{c|}{\textbf{COHFACE}} & \multicolumn{3}{c|}{\textbf{PURE}} & \multicolumn{3}{c}{\textbf{PURE (MPEG-4)}} \\
\textbf{Method} & \textit{MAE}$\downarrow$ & \textit{RMSE}$\downarrow$ & \textit{R}$\uparrow$ & \textit{MAE}$\downarrow$ & \textit{RMSE}$\downarrow$ & \textit{R}$\uparrow$ & \textit{MAE}$\downarrow$ & \textit{RMSE}$\downarrow$ & \textit{R}$\uparrow$\\ 
 \hline\hline
 CHROM \cite{chrom} & 7.80 & 12.45 & 0.26 & 2.07 & 2.50 & 0.99 & 6.29 & 11.36 & 0.55\\
 2SR \cite{ssr} & 20.98 & 25.84 & -0.32 & 2.44 & 3.06 & 0.98 & 5.78 & 12.81 & 0.98\\ 
 POS \cite{pos} & 13.43 & 17.05 & 0.07 & 3.14 & 10.57 & 0.95 & - & - & - \\ \hline
 HR-CNN \cite{HRCNN} & 8.10 & 10.78 & 0.29 & 1.84 & 2.37 & 0.98 & 8.72 & 11.00 & 0.70\\ 
 DeepPhys \cite{deepphys} & 6.89 & 13.89  & 0.34 & 0.83 & 1.54 & 0.99 & 3.10 & 9.37 & 0.83\\ 
 PhysNet \cite{phys} & 8.59 & 11.60  & 0.36 & 1.90 & 3.44 & 0.98 & 5.39 & 11.05 & 0.76\\
 CNN+ConvLSTM \cite{convlstm} & 7.31 & 11.88 & 0.36 & 0.88 & 1.58 & 0.99 & - & - & - \\
 DeeprPPG \cite{deep} & 3.07 & 7.06  & 0.86 & \underline{0.28} & \textbf{0.43} & 0.99 & - & - & -\\
 VitaSi \cite{vitasi} & 7.16 & 9.59 & 0.61 & - & - & - & - & - & -  \\
 MultiHeirCNN \cite{heir} & 5.57 & 9.50 & 0.75 & - & - & - & - & - & - \\
 ETA-rPPGNet \cite{eta} & 4.67 & 6.65  & 0.77 & 0.34  & 0.77 & 0.99 & 2.66 & 6.48 & 0.92\\
 CNN+Att. \cite{strip} & 5.19 & 7.52 & 0.68 & 0.74 & 1.21 & \textbf{1.00} & - & - & -\\ 

 POS+MOT+CNN \cite{hurppg} &  - & - & - & \textbf{0.23} & \underline{0.48} & 0.99 & - & - & - \\
  CDC-CAN \cite{cdt} & \underline{1.71} & \textbf{3.57} & \textbf{0.96} & 0.78 & 1.07 & 0.99 & - & - & -\\ 
 CDCA-rPPGNet \cite{cdcrppg} &  - & - & - & 0.46 & 0.90 & 0.99 & - & - & - \\
 InstTrans \cite{insttrans} & 19.66 & 22.65 & - & - & - & - & - & - & -\\
 RADIANT \cite{radiant} & 8.01 & 10.12 & - & - & - & - & - & - & -\\
  DemodFormer \cite{demodulation} & 4.78 & 7.06 & 0.86 & 1.53 & 2.29 & 0.99 & - & - & -\\ 
CDCCA-RPPGFormer \cite{cdcca} & - & - & - & 0.41 & 0.66 & 0.99 & - & - & -\\
Dual-TokenLearner \cite{dualpath} & - & - & - & 0.37 & 0.68 & 0.99 & - & - & -\\

  TFA-PFE \cite{aaai_tfa} & \textbf{1.31} & 3.92 & - & 1.44 & 2.50 & - & - & - & - \\
  \hline
Supervised baseline (3D) & 2.62 & 4.59 & 0.90 & 0.47 & 0.58 & 0.99 & 0.97 & 1.2 & 0.99\\ 
Supervised baseline ((2+1)D) & 2.68 & 4.42 & 0.90 & 0.50 & 0.62 & 0.99 & 1.06 & 1.52 & 0.99\\ 
  \hline 
  Ours (Self-supervised) w/o neg. & 2.45 & 4.25  & 0.92 & 0.46 & 0.58 & 0.99 & \underline{0.78} & \underline{0.95} & 0.99\\ 
Ours (Self-supervised)  & 2.16 & \underline{3.61}  & \underline{0.94} & 0.43 & 0.58 & 0.99 & \textbf{0.74} & \textbf{0.93} & 0.99\\ 
 \hline
\end{tabular}
\label{tab:complete}
\end{table*}

\begin{figure*}
\centering
\small
\begin{tabular}{cc}
  \includegraphics[width=92mm]{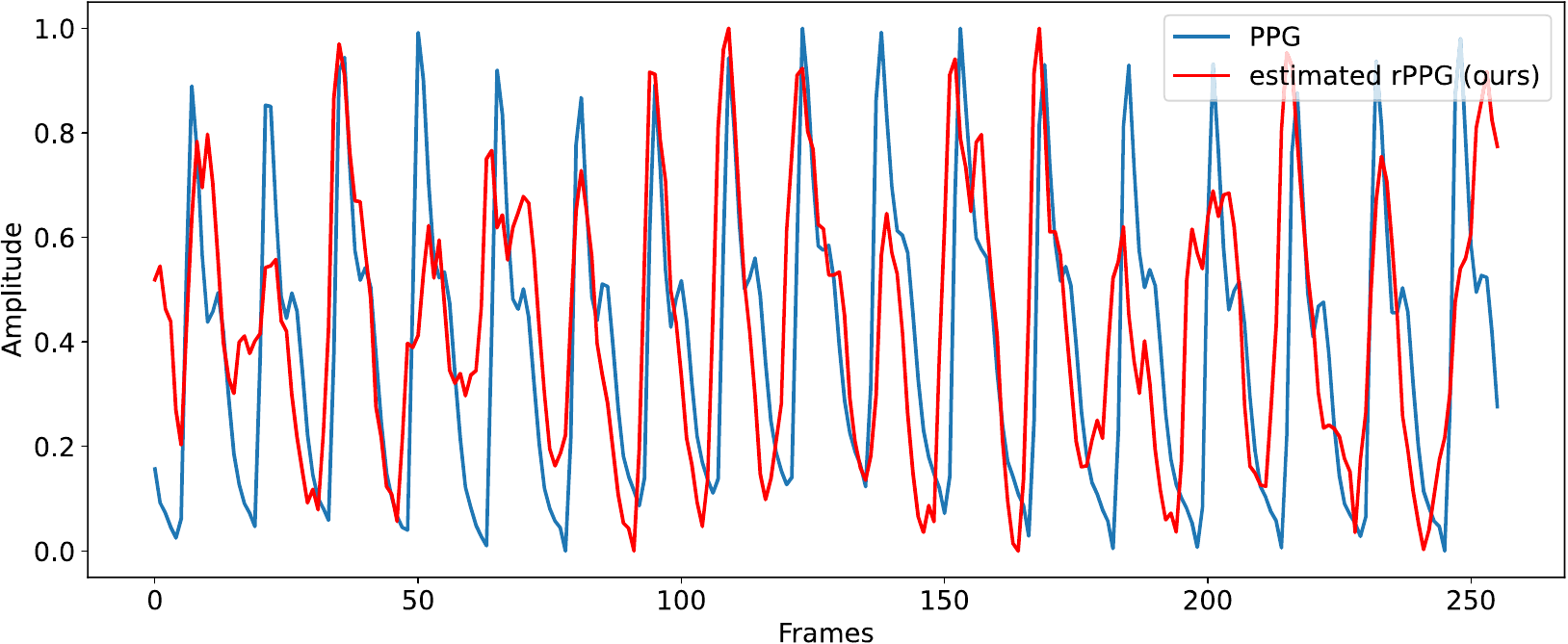} & \includegraphics[width=92mm]{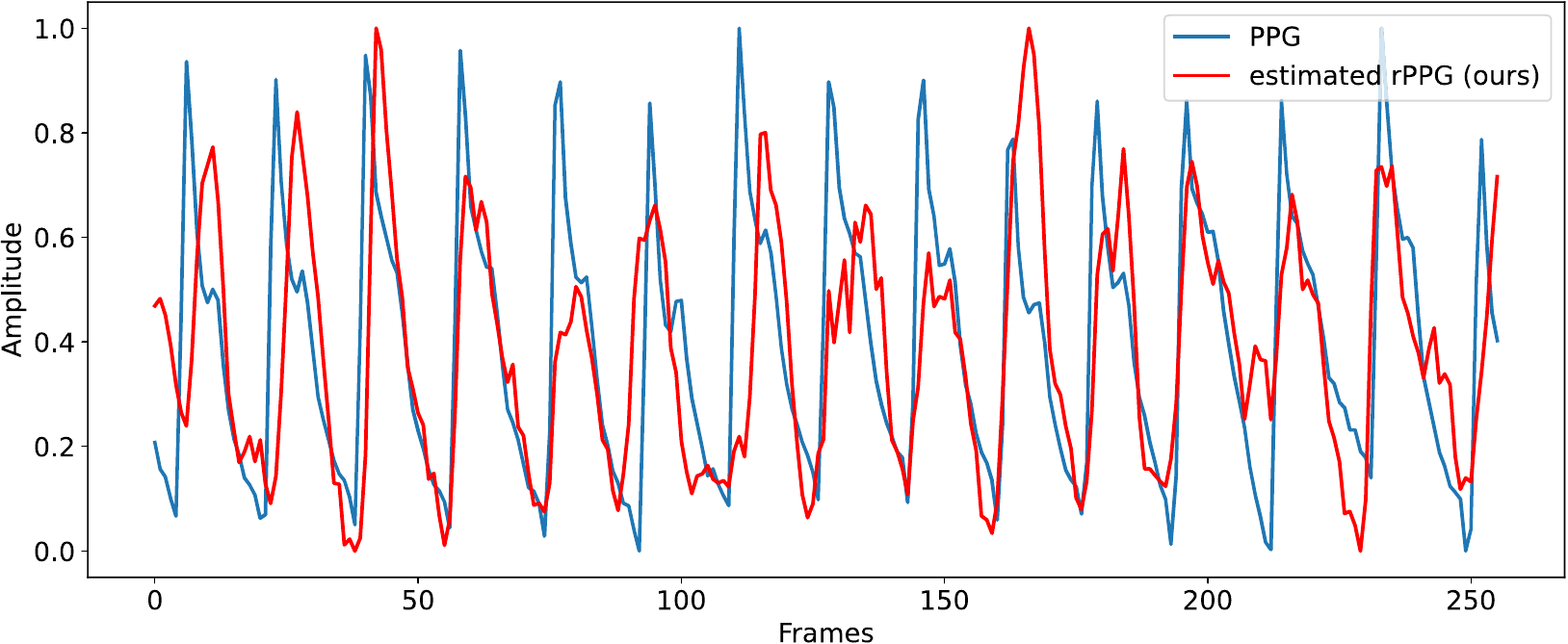} \\
  a) COHFACE - artificial illumination & b) COHFACE - natural illumination \\ \\
  \includegraphics[width=92mm]{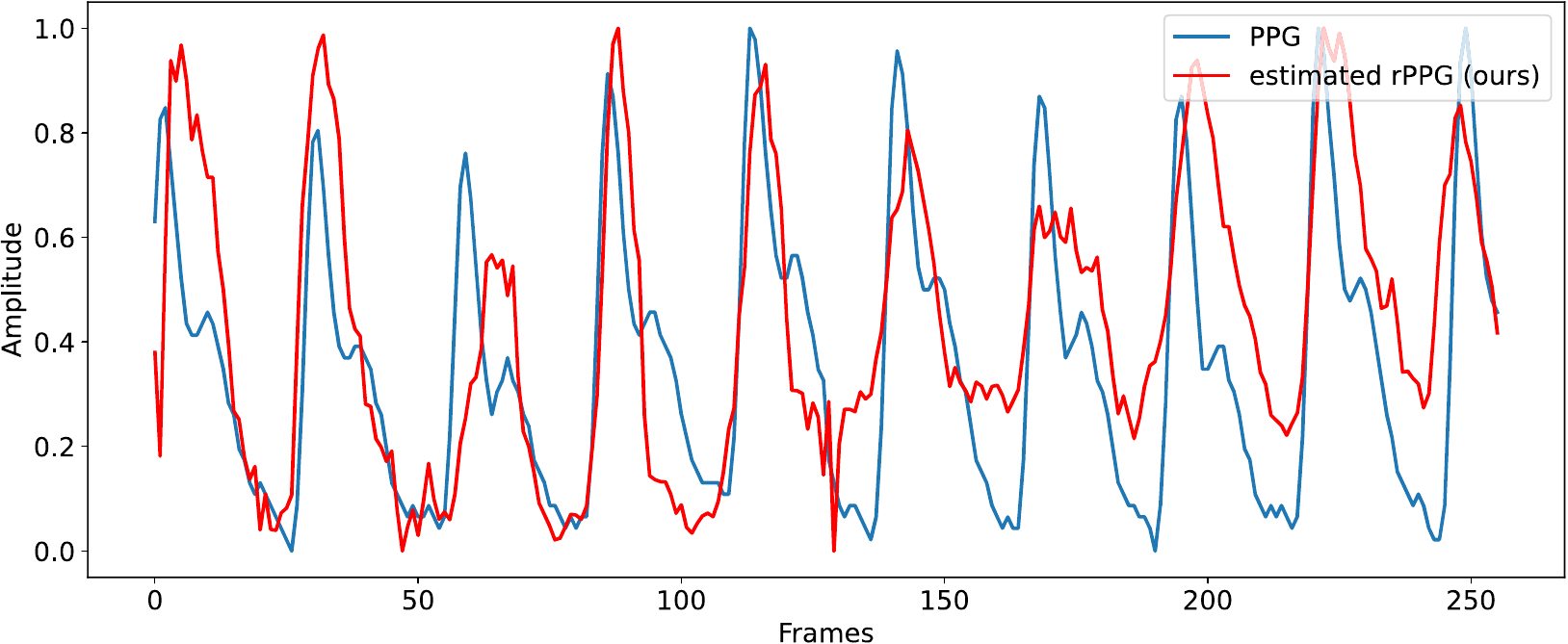} & \includegraphics[width=92mm]{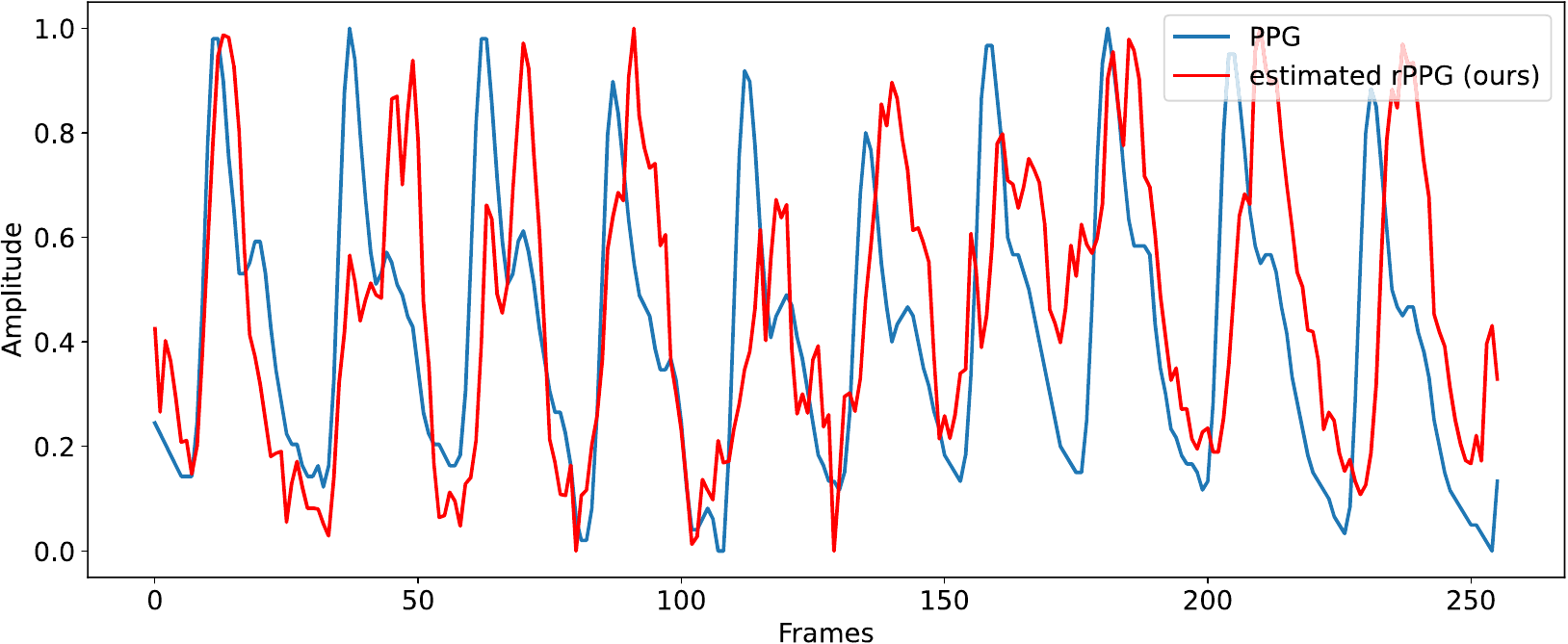} \\
  c) PURE (MPEG-4) - still face & d) PURE (MPEG-4) - talking face  \\ \\
  \includegraphics[width=92mm]{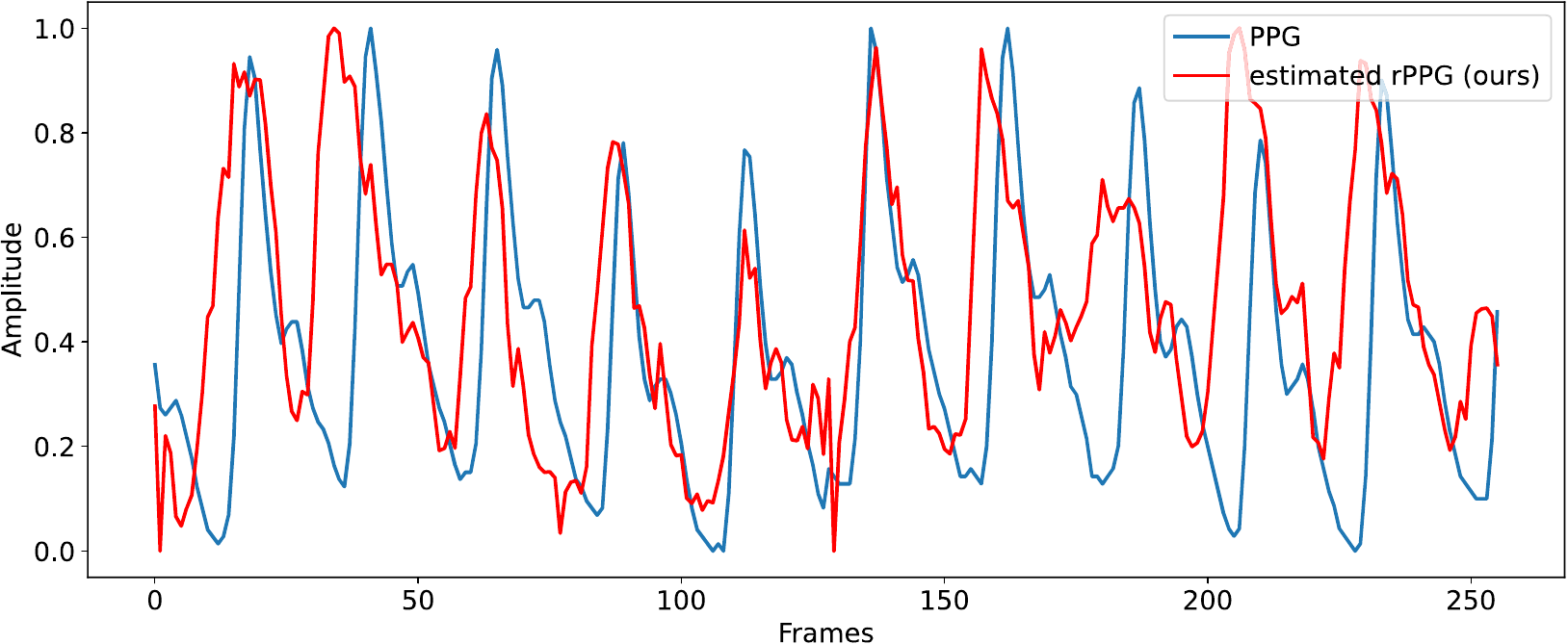} & \includegraphics[width=92mm]{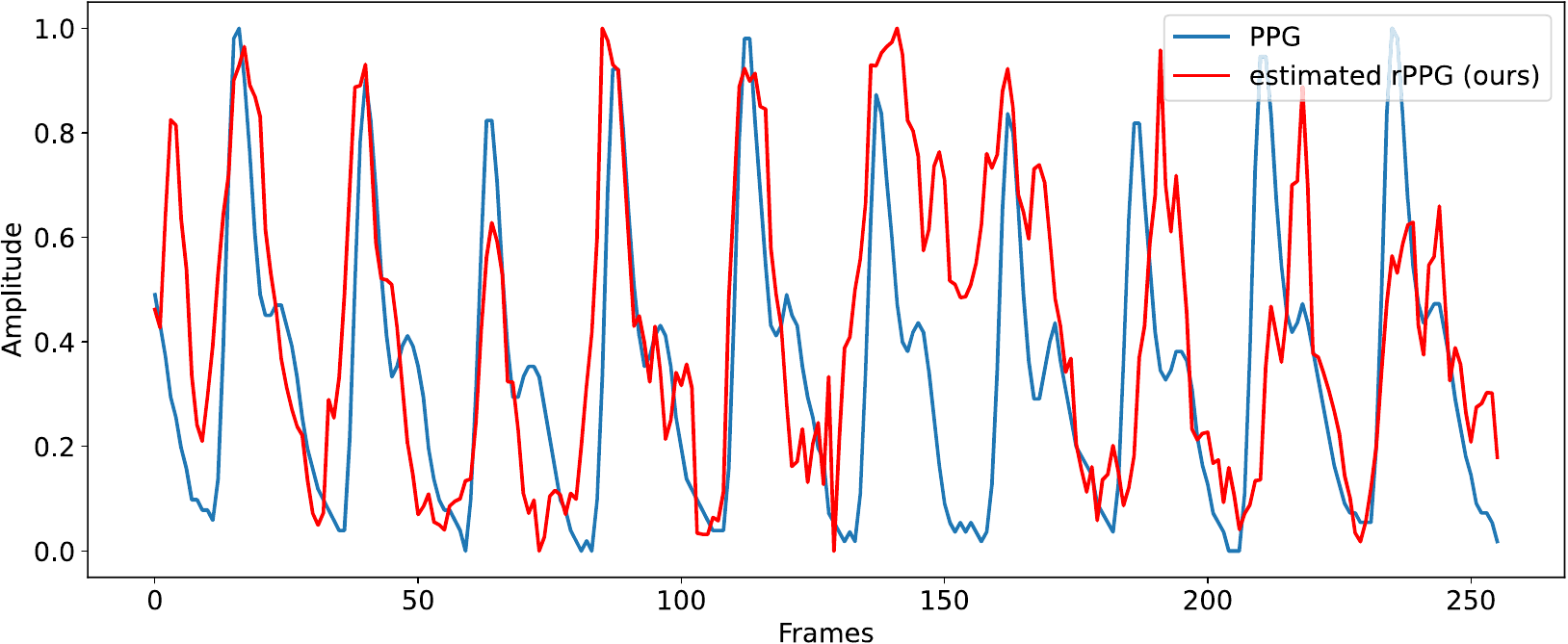} \\
  e) PURE (MPEG-4) - slow head translation & f) PURE (MPEG-4) - fast head translation  \\ \\
  \includegraphics[width=92mm]{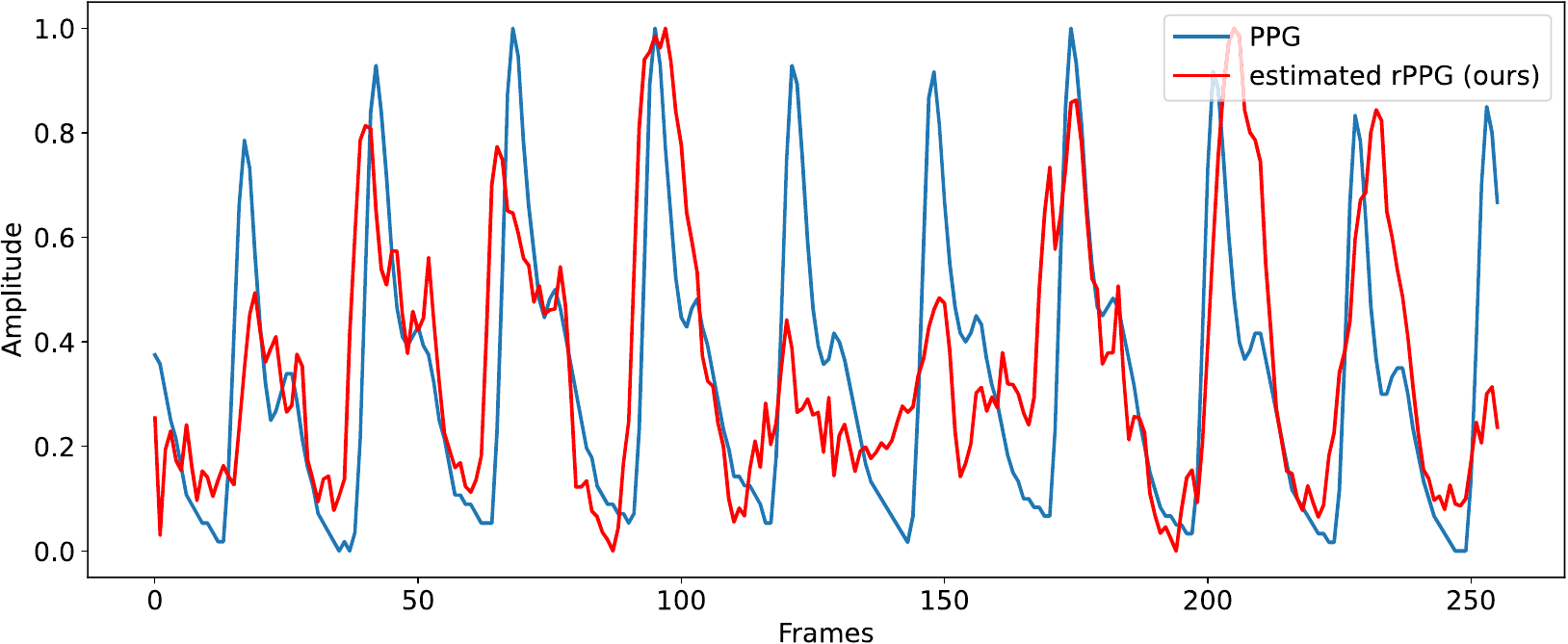} & \includegraphics[width=92mm]{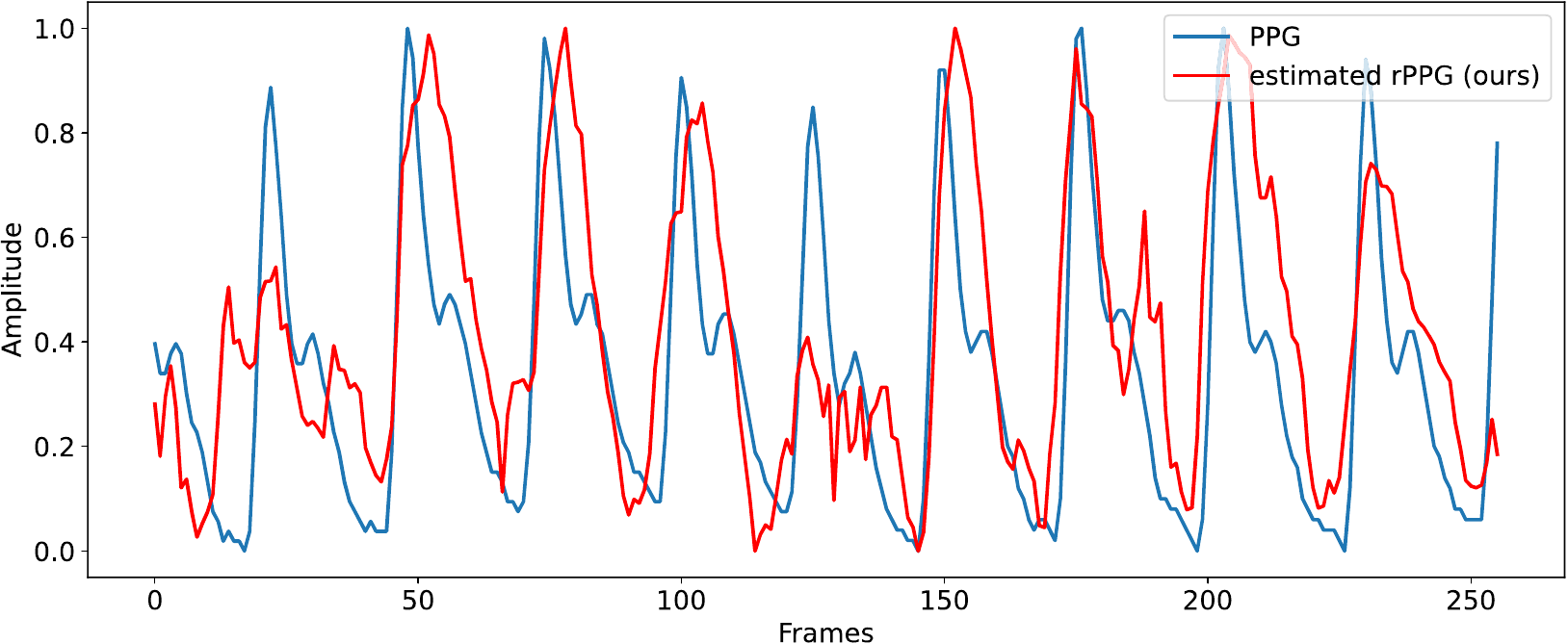} \\
  g) PURE (MPEG-4) - slow head rotation & h) PURE (MPEG-4) - medium head rotation  \\ \\
\end{tabular}
\caption{Visualization of predicted rPPG for different conditions presented in the datasets. We observe that the peaks of the rPPG signals are highly aligned with the peaks of the corresponding PPG signals.}
\label{fig:samplesig}
\end{figure*}

\section{Results}

In this section, we present and discuss our results. We first compare the results with the existing prior works described above. Thereafter we study the impact of using a different technique for video representation learning and the impact of using negative pairs for self-supervised pre-training. Moreover, we study the effects of using different RoIs, namely the combined and individual regions of the forehead and the cheek, and also the different augmentations for self-supervised pre-training. Lastly, we also compare the performance of the supervised and the proposed self-supervised methods on reduced amounts of labeled data.

\subsection{Performance and comparison}
Table \ref{tab:complete} presents the results of our method on COHFACE and PURE, in comparison to prior works. The results show that our method approaches the state-of-the-art on both datasets with respect to all three evaluation metrics. A number of prior works \cite{HRCNN, eta} have additionally used a compressed version of the PURE dataset in MPEG-4 Visual format, which is denoted by `PURE (MPEG-4)'. We also use this approach for a more thorough evaluation of our solution, given that this compression is lossy, meaning that the quality of the data will decrease. We observe that on this dataset, our method achieves superior results compared to other works, indicating low sensitivity with respect to data quality. Overall, we achieved strong results compared to prior works, especially the recent and more advanced methods. We attribute this boost in performance to a number of components of our proposed pipeline namely the RoI detection (to detect and process only the significant facial skin regions instead of the entire face), the self-supervised contrastive pre-training (to learn robust representations), and finally the use of smooth L1 loss in the fine-tuning stage (to combine both L1 and L2 losses for smoother optimization).

\begin{figure}[!t]
\centering
    \includegraphics[width=1\columnwidth]{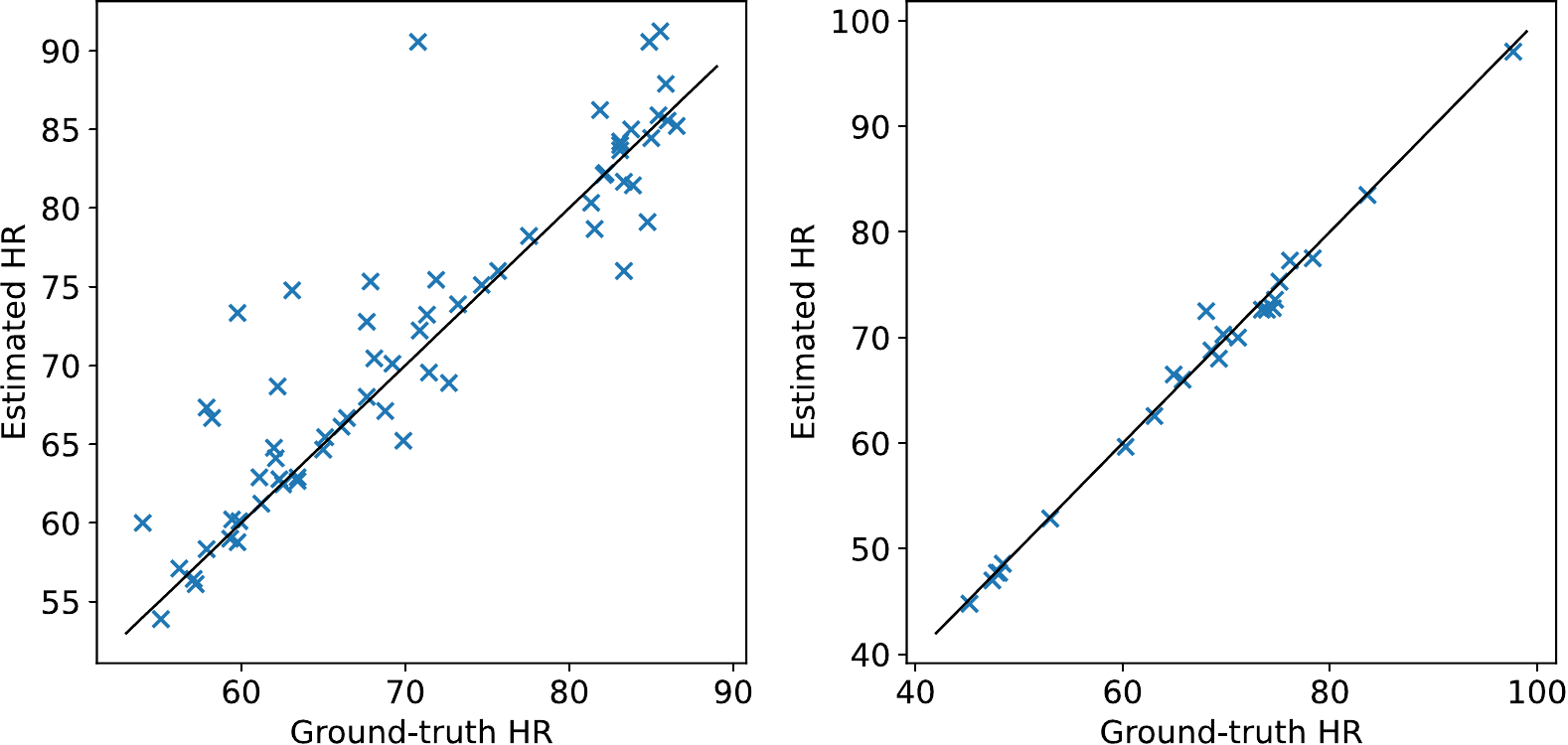}
\caption{Correlation plots for COHFACE (left), and PURE (MPEG-4) (right). The plots show that our results correlate well with the ground-truth.}
\label{fig:corr}
\end{figure}

\begin{figure}[!t]
\centering
    \includegraphics[width=1\columnwidth]{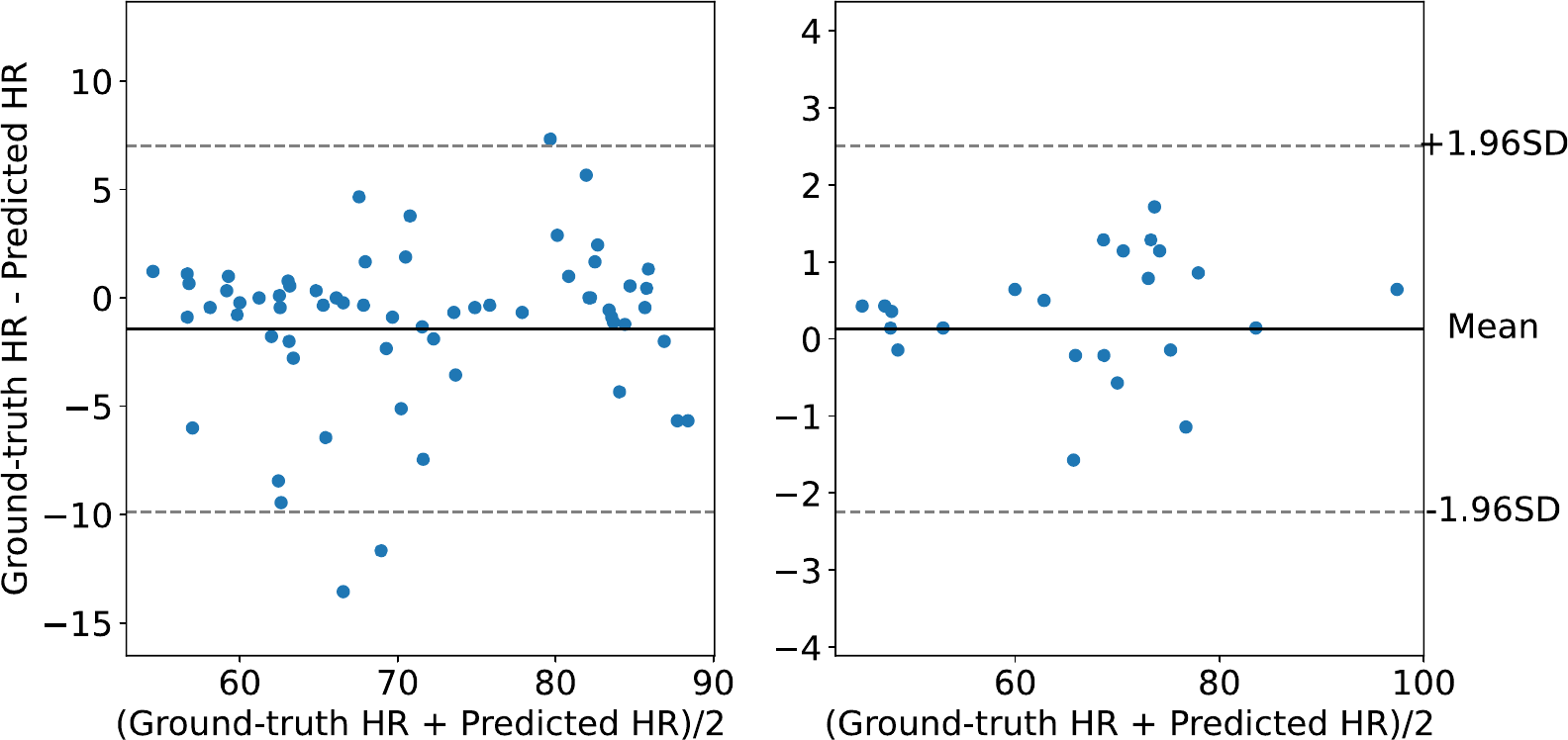}
\caption{B\&A plots for COHFACE (left), and PURE (MPEG-4) (right). The plots depict that our results lie within the limits of agreement.}
\label{fig:bland}
\end{figure}

Additionally, we implement two supervised versions of our model, one using the same 3D convolutions used in our final solution, while in the other, we use (2+1)D convolutions. These baselines are devised by initializing the encoder weights randomly and training them only using the labeled data in a fully supervised manner. We observe that our proposed method outperforms all the baselines by considerable margins on COHFACE and with smaller margins on PURE, demonstrating the clear benefits of the self-supervised aspect of our approach. 
This stands for both cases of using as well as not using the negative pairs for the self-supervised pre-training. To gain a better understanding of the quality of the rPPG signals produced by our model, we visualize sample segments of the estimated rPPG signals along with the ground-truth PPG signals in Figure \ref{fig:samplesig} for varying conditions posed in the datasets. We observe that our model produces high-quality rPPG signals, especially with the peaks being highly aligned with the corresponding PPG signals, which is the key factor in measuring metrics such as HR and heart rate variability. We also explore the correlation and the Bland-Altman (B\&A) plots to better visualize the relation between our results and the ground-truth in Figures \ref{fig:corr} and \ref{fig:bland}. As can be seen in the correlation plots between the predicted and the ground-truth HR values, our results correlate well with the ground-truth values with very few outliers. Similarly in the B\&A plots, our results generally lie within the limits of agreement for both datasets.

Since PURE is stored in lossless format, our work and several prior works obtain an MAE of less than 1, and in some cases even less than 0.5. Moreover, the improvement of self-supervised training over fully-supervised training is minimal (less than 0.1 in MAE). However, we notice that there is considerable improvement in self-supervised learning over the supervised method when using PURE (MPEG-4) instead of PURE. To better study the effects of using self-supervised learning over supervised learning and to further evaluate the performance of our proposed solution with respect to artifacts introduced through video compression \cite{vid_artifact}, we only use PURE (MPEG-4) along with COHFACE for all the subsequent experiments. 

\begin{table*}[!t]
\centering
\caption{Impact of different encoders for pre-training (full RoI).}
\small
\setlength
\tabcolsep{6pt}
\begin{tabular}{l|c c c|c c c|c c c|c c c}
\hline
    & \multicolumn{6}{c|} {COHFACE} & \multicolumn{6}{c} {PURE (MPEG-4)}\\
 & \multicolumn{3}{c}{\textbf{Encoder A}} & \multicolumn{3}{c|}{\textbf{Encoder B }}  & \multicolumn{3}{c|}{\textbf{Encoder A}} & \multicolumn{3}{c}{\textbf{Encoder B }}\\
\textbf{Augmentation} & \textit{MAE}$\downarrow$ & \textit{RMSE}$\downarrow$ & \textit{R}$\uparrow$ & \textit{MAE}$\downarrow$ & \textit{RMSE}$\downarrow$ & \textit{R}$\uparrow$  & \textit{MAE}$\downarrow$ & \textit{RMSE}$\downarrow$ & \textit{R}$\uparrow$ & \textit{MAE}$\downarrow$ & \textit{RMSE}$\downarrow$ & \textit{R}$\uparrow$\\ 
 \hline\hline
Crop & 2.96 & 4.44 & 0.90 &  3.09 & 5.58 & 0.86 & 0.83 & 1.22 & 0.99 & 1.11 & 1.43 & 0.99\\ 
 Rot & 2.78 & 4.84 & 0.88 &  \textbf{2.14} & 3.61 & \textbf{0.94} & 0.81 & 1.05 & 0.99 & 1.25 & 1.82 & 0.99\\
 Flip & \textbf{2.16} & \textbf{3.61} & \textbf{0.94} & 2.57 & 4.08 & 0.92 & 0.88 & 1.20 & 0.99&  1.10 & 1.60 & 0.99\\ \hline
 Reverse & 2.51 & 3.98 & 0.93 &  2.18 & \textbf{3.50} & 0.94 & 0.96 & 1.28 & 0.99 & \textbf{0.72} & \textbf{1.02} & 0.99\\ 
 Reorder & 2.59 & 4.32 & 0.91 &  2.48 & 4.07 & 0.91 & 1.18 & 1.79 & 0.99 & 1.46 & 2.81 & 0.97\\
 Shuffle & 2.22 & 3.67 & 0.94 & 2.40 & 3.68 & 0.93 & \textbf{0.74} & \textbf{0.93} & 0.99& 1.58 & 2.92 & 0.97\\ \hline
 Supervised baseline & 2.62 & 4.59 & 0.90 & 2.68 & 4.42 & 0.90 & 0.97 & 1.20 & 0.99 & 1.06 & 1.52 & 0.99\\ 
  \hline
\end{tabular}
\label{allenc}
\end{table*}

\begin{table*}[!t]
\centering
\caption{Impact of different encoders for pre-training (cheek as RoI).}
\small
\setlength
\tabcolsep{6pt}
\begin{tabular}{l|c c c|c c c|c c c|c c c}
\hline
& \multicolumn{6}{c|} {COHFACE} & \multicolumn{6}{c} {PURE (MPEG-4)}\\
 & \multicolumn{3}{c}{\textbf{Encoder A}} & \multicolumn{3}{c|}{\textbf{Encoder B }}  & \multicolumn{3}{c|}{\textbf{Encoder A}} & \multicolumn{3}{c}{\textbf{Encoder B }}\\
\textbf{Augmentation} & \textit{MAE}$\downarrow$ & \textit{RMSE}$\downarrow$ & \textit{R}$\uparrow$ & \textit{MAE}$\downarrow$ & \textit{RMSE}$\downarrow$ & \textit{R}$\uparrow$  & \textit{MAE}$\downarrow$ & \textit{RMSE}$\downarrow$ & \textit{R}$\uparrow$ & \textit{MAE}$\downarrow$ & \textit{RMSE}$\downarrow$ & \textit{R}$\uparrow$\\ 
 \hline\hline
Crop & 3.66 & 6.09 & 0.83 &  3.57 & 5.56 & 0.84 & 1.93 & 3.00 & 0.97 & 1.65 & \textbf{2.23} & 0.98\\ 
 Rot & \textbf{2.55} & \textbf{3.92} & \textbf{0.90} & \textbf{2.63} & \textbf{4.02} & \textbf{0.89} & 1.13 & 1.56 & 0.99 &  \textbf{1.52} & 2.41 & 0.98\\
 Flip & 2.78 & 4.71 & 0.89 & 3.63 & 6.04 & 0.85 & \textbf{0.89} &\textbf{1.25} & \textbf{0.99} & 1.74 & 2.45 & 0.98\\ \hline
 Reverse & 2.88 & 4.81 & 0.89 & 3.45 & 5.37 & 0.86 & 1.21 &  1.89 & 0.99 & 1.73 & 2.55 & 0.98\\ 
 Reorder & 3.23 & 5.08 & 0.88 & 3.05 & 5.25 & 0.87 & 1.58 & 2.20 & 0.98 & 1.83 & 2.63 & 0.98\\
 Shuffle & 3.14 & 5.17 & 0.87 & 3.12 & 4.70 & 0.88 & 1.48 & 2.10 & 0.98 & 1.99 & 3.18 & 0.97\\
  \hline
 Supervised baseline & 3.03 & 5.17 & 0.87 & 3.28 & 5.31 & 0.85 & 1.46 & 2.64 & 0.98 & 1.84 & 2.44 & 0.98\\ 
 \hline
\end{tabular}
\label{cheekenc}
\end{table*}

\begin{table*}[!t]
\centering
\caption{Impact of different encoders for pre-training (forehead as RoI).}
\small
\setlength
\tabcolsep{6pt}
\begin{tabular}{l|c c c|c c c|c c c|c c c}
\hline
& \multicolumn{6}{c|} {COHFACE} & \multicolumn{6}{c} {PURE (MPEG-4)}\\
 & \multicolumn{3}{c}{\textbf{Encoder A}} & \multicolumn{3}{c|}{\textbf{Encoder B }}  & \multicolumn{3}{c|}{\textbf{Encoder A}} & \multicolumn{3}{c}{\textbf{Encoder B }}\\
\textbf{Augmentation} & \textit{MAE}$\downarrow$ & \textit{RMSE}$\downarrow$ & \textit{R}$\uparrow$ & \textit{MAE}$\downarrow$ & \textit{RMSE}$\downarrow$ & \textit{R}$\uparrow$  & \textit{MAE}$\downarrow$ & \textit{RMSE}$\downarrow$ & \textit{R}$\uparrow$ & \textit{MAE}$\downarrow$ & \textit{RMSE}$\downarrow$ & \textit{R}$\uparrow$\\ 
 \hline\hline
Crop & 4.68 & 7.11 & 0.78 &  4.74 & 7.33 & 0.73 & 2.55 & 3.62 & 0.96 & 3.13  & 4.94 & 0.94\\ 
 Rot & \textbf{4.01} & \textbf{5.55} & \textbf{0.84} &  4.79 & \textbf{7.13} &\textbf{0.80} & 2.23 & 3.37 & 0.97 & 2.12 & 3.37 & 0.96\\
 Flip & 4.67 & 7.22 & 0.75 &  5.00 & 7.50 & 0.72 & 2.55 & 3.36 & 0.97& 2.22 & 3.14 & 0.97\\ \hline
 Reverse & 4.54 & 6.55 & 0.77 & 5.53 & 8.25 & 0.66 & 2.38 & 3.33 & 0.96 &  2.06 & 3.60 & 0.96\\ 
 Reorder & 5.11 & 7.43 & 0.72 &  5.31 & 8.48 & 0.66 & \textbf{1.90 }& \textbf{2.58} & \textbf{0.98} & 4.70 & 8.24 & 0.81\\
 Shuffle & 5.28 & 7.93 & 0.74 &  5.71 & 8.43 & 0.71 & 2.36 & 3.88 & 0.95 &\textbf{1.87} & \textbf{2.90} & \textbf{0.97}\\ 
 \hline
 Supervised baseline & 4.96 & 7.32 & 0.71 &\textbf{4.73} & 7.33 & 0.72 &  2.47 & 4.10 & 0.95 &  2.61 & 4.44 & 0.94\\ 
 \hline
\end{tabular}
\label{forenc}
\end{table*}

\subsection{Impact of 3D convolutions}
In Tables \ref{allenc}, \ref{cheekenc}, and \ref{forenc}, we compare the effect of using different video encoding methods by experimenting with 3D and (2+1)D convolutions. Encoder A refers to the encoder using 3D convolution while Encoder B refers to the encoder using (2+1)D convolutions. In the case of the fully-supervised baselines for all three instances of the RoI, using (2+1)D convolutions gives comparable results to the 3D counterpart. However, among the results obtained after self-supervised pre-training and fine-tuning, the best results across all three metrics were obtained for the 3D convolution-based encoder. Since rPPG estimation relies on very slight changes in skin color, the additional non-linearities brought along with using (2+1)D convolution might interfere with the training process rather than help in some cases. This can be seen especially in the case of PURE (MPEG-4) in Table \ref{cheekenc} wherein the error values obtained using Encoder B in the self-supervised approach are nearly double the values obtained using Encoder A. Moreover, in the case of COHFACE in Table \ref{forenc}, using Encoder B did not give any improvement in terms of MAE when using the self-supervised approach. Nevertheless, there is substantial improvement shown by the self-supervised approach over the supervised approach for both the encoders. For COHFACE, compared to the fully supervised learning baseline for Encoder A, the self-supervised learning approach using the flip augmentation, reduces MAE from 2.62 to 2.16, and RMSE from 4.59 to 3.61, while increasing \textit{R} from 0.90 to 0.94.  For PURE (MPEG-4), compared to the fully supervised learning baseline, the self-supervised learning approach using the shuffle augmentation, reduces MAE from 0.97 to 0.74 and RMSE from 1.20 to 0.93. Likewise, for COHFACE, compared to the fully supervised learning baseline for Encoder B, the self-supervised learning approach using the rotation augmentation,  reduces MAE from 2.68 to 2.14 and RMSE from 4.42 to 3.61, while increasing \textit{R} from 0.90 to 0.94. For PURE (MPEG-4), compared to the fully supervised learning baseline, the self-supervised learning approach using the reverse augmentation, reduces MAE from 1.06 to 0.72 and RMSE from 1.52 to 1.02. We observe similar improvements when using the cheek and the forehead as separate RoIs in Tables \ref{cheekenc} and \ref{forenc}, except in the setting with using the forehead as the RoI along with Encoder B for COHFACE where the MAE are very similar.

\begin{table*}[!t]
\centering
\caption{Impact of including and excluding negative pairs in pre-training (full RoI).}
\small
\setlength
\tabcolsep{6pt}
\begin{tabular}{l|c c c|c c c|c c c|c c c}
\hline
    & \multicolumn{6}{c|} {COHFACE} & \multicolumn{6}{c} {PURE (MPEG-4)}\\
 & \multicolumn{3}{c}{\textbf{With negative pairs}} & \multicolumn{3}{c|}{\textbf{Without  negative pairs}}  & \multicolumn{3}{c|}{\textbf{With negative pairs}} & \multicolumn{3}{c}{\textbf{Without negative pairs}}\\
\textbf{Augmentation} & \textit{MAE}$\downarrow$ & \textit{RMSE}$\downarrow$ & \textit{R}$\uparrow$ & \textit{MAE}$\downarrow$ & \textit{RMSE}$\downarrow$ & \textit{R}$\uparrow$  & \textit{MAE}$\downarrow$ & \textit{RMSE}$\downarrow$ & \textit{R}$\uparrow$ & \textit{MAE}$\downarrow$ & \textit{RMSE}$\downarrow$ & \textit{R}$\uparrow$\\ 
 \hline\hline
Crop & 2.96 & 4.44 & 0.90 &  2.54 & 4.68 & 0.89 & 0.83 & 1.22 & 0.99 & \textbf{0.78 } &\textbf{ 0.95} & 0.99\\ 
 Rot & 2.78 & 4.84 & 0.88 &  2.78 & 4.51 & 0.91 & 0.81 & 1.05 & 0.99 & 0.99 & 1.35 & 0.99 \\
 Flip & \textbf{2.16} & \textbf{3.61} & \textbf{0.94} & \textbf{2.45} & \textbf{4.25} & \textbf{0.92} & 0.88 & 1.20 & 0.99&  0.92 & 1.13 & 0.99\\ \hline
 Reverse & 2.51 & 3.98 & 0.93 & 2.76 & 4.56 & 0.90 & 0.96 & 1.28 & 0.99 & 0.84 & 1.05 & 0.99 \\ 
 Reorder & 2.59 & 4.32 & 0.91 &  2.59 & 4.28 & 0.91 & 1.18 & 1.79 & 0.99 &  1.09 & 1.85 & 0.99\\
 Shuffle & 2.22 & 3.67 & 0.94 & 2.82 & 5.26 & 0.88 & \textbf{0.74} & \textbf{0.93} & 0.99& 0.93 & 1.40 & 0.99\\ 
 \hline
 Supervised baseline & 2.62 & 4.59 & 0.90 & 2.62 & 4.59 & 0.90 & 0.97 & 1.20 & 0.99 & 0.97 & 1.20 & 0.99\\ 
  \hline
\end{tabular}
\label{allneg}
\end{table*}

\begin{table*}[!t]
\centering
\caption{Impact of including and excluding negative pairs in pre-training (cheek as RoI).}
\small
\setlength
\tabcolsep{6pt}
\begin{tabular}{l|c c c|c c c|c c c|c c c}
\hline
& \multicolumn{6}{c|} {COHFACE} & \multicolumn{6}{c} {PURE (MPEG-4)}\\
 & \multicolumn{3}{c}{\textbf{With negative pairs}} & \multicolumn{3}{c|}{\textbf{Without  negative pairs}}  & \multicolumn{3}{c|}{\textbf{With negative pairs}} & \multicolumn{3}{c}{\textbf{Without negative pairs}}\\
\textbf{Augmentation} & \textit{MAE}$\downarrow$ & \textit{RMSE}$\downarrow$ & \textit{R}$\uparrow$ & \textit{MAE}$\downarrow$ & \textit{RMSE}$\downarrow$ & \textit{R}$\uparrow$  & \textit{MAE}$\downarrow$ & \textit{RMSE}$\downarrow$ & \textit{R}$\uparrow$ & \textit{MAE}$\downarrow$ & \textit{RMSE}$\downarrow$ & \textit{R}$\uparrow$\\ 
 \hline\hline
Crop & 3.66 & 6.09 & 0.83 &  2.87 & 4.96 & 0.87 & 1.93 & 3.00 & 0.97 &  1.27 & 1.65 & 0.99\\ 
 Rot & \textbf{2.55} & \textbf{3.92} & \textbf{0.90} & 3.38 & 5.29 & 0.86 & 1.13 & 1.56 & 0.99 & \textbf{0.93} &\textbf{1.29 }& 0.99\\
 Flip & 2.78 & 4.71 & 0.89 & 3.27 & 5.58 & 0.84 & \textbf{0.89} &\textbf{1.25} & 0.99 & 1.47 & 2.09 &  0.99 \\ \hline
 Reverse & 2.88 & 4.81 & 0.89 &  \textbf{2.75} & \textbf{4.42} & \textbf{0.90} &  1.21 & 1.89 &  0.99 & 1.66 & 2.68 & 0.98\\
 Reorder & 3.23 & 5.08 & 0.88 & 2.93 & 4.61 & 0.89   & 1.58 & 2.20 & 0.98 &  1.60 & 2.50 &  0.98 \\
 Shuffle & 3.14 & 5.17 & 0.87 & 3.01 & 4.95 & 0.87 & 1.48 & 2.10 & 0.98 &  1.18 & 1.70 & 0.99 \\
  \hline
 Supervised baseline & 3.03 & 5.17 & 0.87 & 3.03 & 5.17 & 0.87 & 1.46 & 2.64 & 0.98 & 1.46 & 2.64 & 0.98 \\ 
 \hline
\end{tabular}
\label{cheekneg}
\end{table*}

\begin{table*}[!t]
\centering
\caption{Impact of including and excluding negative pairs in pre-training (forehead as RoI).}
\small
\setlength
\tabcolsep{6pt}
\begin{tabular}{l|c c c|c c c|c c c|c c c}
\hline
& \multicolumn{6}{c|} {COHFACE} & \multicolumn{6}{c} {PURE (MPEG-4)}\\
 & \multicolumn{3}{c}{\textbf{With negative pairs}} & \multicolumn{3}{c|}{\textbf{Without negative pairs}}  & \multicolumn{3}{c|}{\textbf{With negative pairs}} & \multicolumn{3}{c}{\textbf{Without negative pairs}}\\
\textbf{Augmentation} & \textit{MAE}$\downarrow$ & \textit{RMSE}$\downarrow$ & \textit{R}$\uparrow$ & \textit{MAE}$\downarrow$ & \textit{RMSE}$\downarrow$ & \textit{R}$\uparrow$  & \textit{MAE}$\downarrow$ & \textit{RMSE}$\downarrow$ & \textit{R}$\uparrow$ & \textit{MAE}$\downarrow$ & \textit{RMSE}$\downarrow$ & \textit{R}$\uparrow$\\ 
 \hline\hline
Crop & 4.68 & 7.11 & 0.78 &  4.89 & 7.24 & 0.76 & 2.55 & 3.62 & 0.96 &  1.91 & \textbf{2.71} & \textbf{0.98} \\ 
Rot & \textbf{4.01} & \textbf{5.55} & \textbf{0.84} & \textbf{4.46} & \textbf{6.65} & \textbf{0.77} & 2.23 & 3.37 & 0.97 & \textbf{1.88} & 2.75 & 0.97 \\
 Flip & 4.67 & 7.22 & 0.75 &  5.05 & 7.12 & 0.76 & 2.55 & 3.36 & 0.97&   2.50 & 3.41 & 0.97 \\ \hline
 Reverse & 4.54 & 6.55 & 0.77 & 4.82 & 7.04 & 0.74  & 2.38 & 3.33 & 0.96 &  2.35 & 3.37 & 0.96 \\ 
 Reorder & 5.11 & 7.43 & 0.72 &  4.89 & 7.86 & 0.73  & \textbf{1.90 }& \textbf{2.58} & \textbf{0.98} & 2.48 & 3.77 & 0.96 \\
 Shuffle & 5.28 & 7.93 & 0.74 & 5.00&7.66& 0.73 & 2.36 & 3.88 & 0.95 & 2.23 & 3.28 & 0.97\\ 
 \hline 
 Supervised baseline & 4.96 & 7.32 & 0.71 & 4.96 & 7.32 & 0.71 &  2.47 & 4.10 & 0.95 &   2.47 & 4.10 & 0.95 \\ 
 \hline
\end{tabular}
\label{foreneg}
\end{table*}

\subsection{Impact of negative pairs in pre-training}
Next, in Tables \ref{allneg}, \ref{cheekneg}, \ref{foreneg}, we compare the effect of using negative pairs for self-supervised pre-training. We observe that in some cases not using the negative pairs for this purpose yields better results in comparison to when the negative pairs are used. However, the best results for almost all RoIs for both datasets are obtained when using the negative pairs for the pre-training. The one exception is the case of using the forehead alone as the RoI for PURE (MPEG-4) wherein the MAE are very similar. In other cases, although not using the negative pairs does not give the best results, it still gives improvement over the fully-supervised baselines. For COHFACE, while processing the combined RoI, using the self-supervised learning approach without negative pairs with the flip augmentation, MAE is reduced from 2.62 to 2.45 and RMSE from 4.59 to 4.25, while increasing \textit{R} from 0.90 to 0.92. Likewise for PURE (MPEG-4), using the crop augmentation reduces MAE from 0.97 to 0.78 and RMSE from 1.20 to 0.95. A similar trend can be observed when using the cheek and the forehead as individual inputs. Overall, we observe in this experiment that the use of negative pairs generally benefits our solution. This can be due to the fact that for the problem of rPPG estimation, it is highly unlikely that two input clips would have identical PPG patterns, or in other words, identical labels (given that we have a regression problem). Thereby when we use negative pairs in our setup, the network essentially learns to distinguish even between those clips that might have some spatial similarities or even similar HR values, yet still different PPG patterns. This will allow for more effective representations to be learned to achieve better overall performance.

\subsection{Impact of different facial regions}
Here, we study the effects of using different facial regions for rPPG estimation. To do so, we use different regions of the face, namely cheeks, forehead, and a combination of both (our original solution), as input to our model. Since the input dimensions will differ when the cheeks or the forehead alone are used, we perform spatial interpolation to scale the input to 64$\times$64 pixels (our original input spatial dimension). Tables \ref{cheekenc} and \ref{forenc} present the results where we observe (considering Encoder A) that when using only the cheek, we obtain results closer to the ones obtained when using the combined RoI of the cheek and forehead. However, this is not the case when we use the forehead as the sole input. This can be primarily due to the forehead region being partially occluded by hair, having wrinkles, and other artifacts. Moreover, we observe that there is significant improvement while using the self-supervised pre-training over fully-supervised baselines when we use the facial regions separately. For COHFACE, when using cheek as the RoI, using the self-supervised learning approach with rotation augmentation, reduced MAE from 3.03 to 2.55, RMSE from 5.17 to 3.92, and increased \textit{R} from 0.87 to 0.90 while for the forehead, the self-supervised learning approach using rotation augmentation, reduced the MAE from 4.96 to 4.01, RMSE from 7.32 to 5.55, and increased \textit{R} from 0.71 to 0.84. Likewise for PURE (MPEG-4), when using cheek as the RoI, using the self-supervised learning approach with flip augmentation, reduced the MAE from 1.46 to 0.89, RMSE from 2.64 to 1.25, and increased \textit{R} from 0.98 to 0.99 and while for the forehead, the self-supervised learning approach using reorder augmentation, reduced the MAE from 2.47 to 1.90, RMSE from 4.10 to 2.58, and increased \textit{R} from 0.95 to 0.98. Nevertheless, there are considerable differences in the values of the metrics obtained whilst using the cheek and the forehead as standalone RoIs. A similar trend can be observed in Tables \ref{cheekneg}, \ref{foreneg}, wherein we do not use the negative pairs of the input RoI in the self-supervised pre-training.

\subsection{Impact of different augmentations}
Next, we explore the impact of different augmentations. Here, we only consider our best setups, i.e., Encoder A with negative pairs using the combined RoI. Revisiting Table \ref{allenc}, we notice that for COHFACE, flip augmentation yields the best results, while for PURE (MPEG-4), shuffle results in the best performance. However, we observe that with the exception of flip, the temporal augmentations gave better results than the spatial augmentations for COHFACE. Similarly in PURE (MPEG-4), with the exception of shuffle, the spatial augmentations provide better results than the temporal ones. We should note that the subjects in the COHFACE dataset have fewer spatial variations compared to the subjects in PURE, as the subjects in PURE were recorded with varying facial movements while the subjects in COHFACE were stationary. When using contrastive learning, robust feature representations are learned which make the model invariant to the augmentation used to generate the pairs \cite{transfer}, which might be the reason for the better overall performance of spatial augmentations in PURE (MPEG-4). Nevertheless, the best performances are given by a spatial augmentation in COHFACE, and a temporal augmentation in the case of PURE (MPEG-4), demonstrating the need for exploration of a wide variety of augmentations for use in contrastive learning \cite{sahar}. 

\subsection{Performance on reduced labels}
\begin{figure}
\centering
    \includegraphics[width=1\columnwidth]{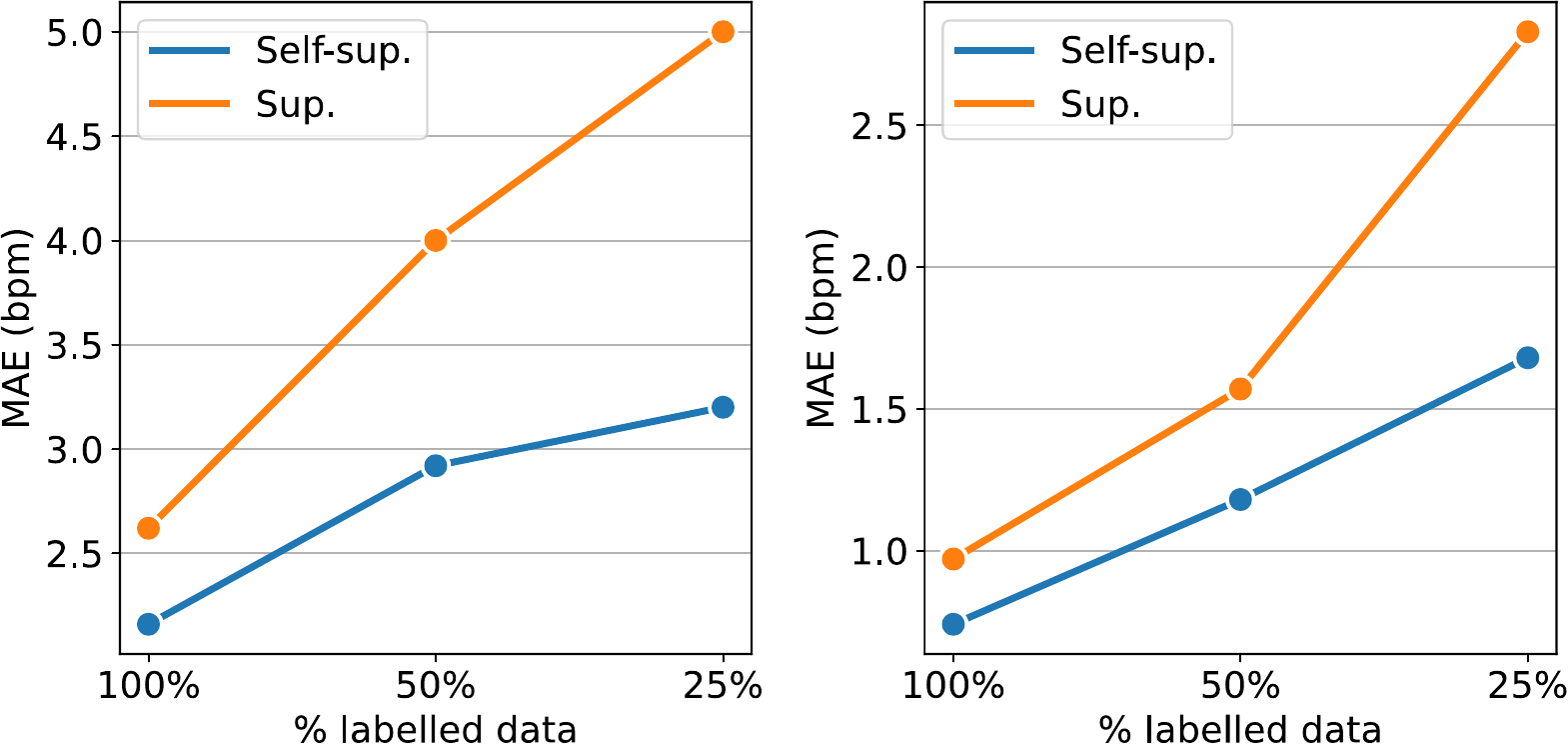}
\caption{Performance of self-supervised and fully supervised approaches on reduced amounts of labelled data for COHFACE (left), and PURE (MPEG-4) (right). The self-supervised approach suffers from lower drops in MAE than the fully supervised approach, showcasing its robustness.}
\label{fig:reduced}
\end{figure}

Lastly, to further illustrate the advantage of using self-supervised vs. fully supervised learning, we compare the two approaches on reduced amounts of labeled data. We first train the encoder (Encoder A) on all the video clips of the training data through contrastive learning as done in all of our previous experiments. Next, for fine-tuning, we randomly select 50\% and 25\% of the video clips and their corresponding PPG signals from the training data, and fine-tune the network using the smooth L1 Loss. For the supervised learning method, we train \textit{supervised baseline (3D)} from scratch on the same randomly selected video clips and PPG signals. Figure \ref{fig:reduced} presents the performance on the full and reduced (50\%, 25\%) training sets when the best augmentations are used to train the model. As we observe, the self-supervised approach leads to more robustness (suffers smaller drops in performance) when dealing with reduced labels on both datasets.  We also note that even with 25\% of the labels, we achieve results better than most prior works utilizing 100\% of the labels, thereby highlighting the key advantage of using contrastive self-supervised learning as a pre-training step.

From the above experiments, we observe that choosing the RoI has a significant impact on the results. This is in line with the findings of \cite{roi_analysis, ppgroi} where it was shown that the RoI needs to be selected carefully for better quality signals. Moreover, there was more improvement in performance on the PURE (MPEG-4) dataset compared to COHFACE. While COHFACE has a higher number of subjects that are not moving, half of the videos have uneven lighting as described in Section \ref{datasets}, which impacts rPPG estimation since rPPG signals are estimated from the light reflected from the surface of the skin. Moreover, the detection and cropping of RoI as described in Section \ref{roidet}, seeks to counteract the effect of facial movements for efficient rPPG estimation. These factors combined account for the better performance and improvement seen in PURE (MPEG-4) over COHFACE. Nevertheless, the results of our self-supervised approach are superior to the supervised baselines and several prior works on both datasets, showcasing strong generalization in the presence of various artifacts.

\section{Conclusion and Future Work}
In this work, we proposed a two-stage method based on the use of self-supervised contrastive pre-training and fine-tuning for remote PPG estimation and HR prediction from facial videos. We showcased that introducing contrastive learning as a pre-training measure helps in more robust learning of the network for the downstream task of rPPG estimation. Subsequently, we performed thorough experiments to validate the different design choices of the method such as the video representation learning technique, the pairing strategy, the augmentations used in the self-supervised pre-training, and the different facial regions to be used as inputs for the entire method. Our comprehensive experiments showed that our self-supervised approach outperforms many fully supervised techniques to approach the state-of-the-art, while also being less reliant on output labels during the training stage.  

For future work, the set and combination of different augmentations could be expanded. Moreover, concepts such as attention mechanisms could be added to our encoder to further focus the model on more salient regions of the face. Additionally, loss functions that have been proven more suitable for regression and signal generation problems could be explored. 

Another avenue where future work can be oriented is toward mitigating bias and ensuring fairness through the proposed algorithm. As rPPG relies on the modulation of light intensities reflected from the skin's surface, its performance can be influenced by various factors including but not limited to age, skin complexion, makeup, cultural characteristics, and more, which exhibit variations across diverse demographic groups. Therefore, our method can be extended to specifically address and mitigate these challenges.

Lastly, it is important to note that electrocardiogram (ECG) signals are the gold standard for measuring the various cardiac vitals associated with health monitoring in clinical settings. However, they cannot be measured remotely as done in the case of PPG signals, and require costly contact-based sensors. This leaves open a possible future research direction, where datasets comprising facial videos along with both PPG (to help train the rPPG algorithm) and ECG (clinical reference) signals can be used to derive cardiac vitals from both estimated rPPG and measured ECG signals to analyze the clinical relevance of the estimated rPPG signals.

\section*{Acknowledgment} The authors would like to thank BMO Bank of Montreal and Mitacs for funding this research.

\bibliographystyle{IEEEbib1}
\bibliography{refs}

\end{document}